\documentclass{bmvc2k}
\usepackage{graphicx}
\usepackage{amsmath,amssymb} 
\usepackage{color}
\usepackage{graphicx}
\usepackage{wrapfig}
\usepackage{breakcites}
\usepackage{wasysym}
\usepackage{empheq}
\usepackage{algorithm}
\usepackage{algpseudocode}
\usepackage{caption} 
\definecolor{linksblue}{rgb}{0.16, 0.36, 0.69}
\definecolor{linksorange}{rgb}{0.95, 0.4, 0.}
\hypersetup{colorlinks, linkcolor=linksorange, citecolor=linksblue}

\newcommand{\gate}{\ensuremath{g}}

\usepackage{multirow}
\usepackage{bbm}

\title{Revisiting Single-gated Mixtures of Experts}

\addauthor{Am\'{e}lie Royer}{aroyer@qti.qualcomm.com}{1}
\addauthor{Ilia Karmanov}{ikarmano@qti.qualcomm.com}{1,2}
\addauthor{Andrii Skliar}{askliar@qti.qualcomm.com}{1}
\addauthor{Babak Ehteshami Bejnordi}{behtesha@qti.qualcomm.com}{1}
\addauthor{Tijmen Blankevoort}{tijmen@qti.qualcomm.com}{1}

\addinstitution{
Qualcomm AI Research\\
Science Park 400\\
Amsterdam\\
The Netherlands
}
\addinstitution{
\footnotesize Now working at NVIDIA Switzerland
}

\runninghead{Royer et al}{Revisiting Single-gated Mixtures of Experts}

\begin{document}
\maketitle

\begin{abstract}
Mixture of Experts (MoE) are rising in popularity as a means to train extremely large-scale models, yet allowing for a reasonable computational cost at inference time.
%
Recent state-of-the-art approaches usually assume a large number of experts, and require training all experts jointly, which often lead to training instabilities such as the router collapsing.
In contrast, in this work, we propose to revisit simple single-gate MoE, which allows for more practical training. 
Key to our work are \textbf{(i)} a base model branch acting both as an early-exit and an \textit{ensembling regularization} scheme, 
\textbf{(ii)} a simple and efficient \textit{asynchronous} training  pipeline without router collapse issues, and finally \textbf{(iii)} an automatic \textit{per-sample} clustering-based initialization. 
We show experimentally that the proposed model obtains  efficiency-to-accuracy trade-offs comparable with other more complex MoE, and outperforms non-mixture baselines. This showcases the merits of even a simple single-gate MoE, and motivates further exploration in this area.

\end{abstract}

\section{Introduction}\label{sec:introduction}
Neural networks are designed to extract a fixed set of exhaustive features for any given image. 
However, images exhibit varying levels of complexity, from simple cases such as single objects on a white background to images with clutter and difficult camera angles. 
Treating both of these cases equally can be wasteful from an efficiency perspective: 
This intuition has given rise to very active research in the fields of conditional computing \cite{Bengio2013CC} and early-exiting \cite{Teerapittayanon2016branchynet,Huang2018MultiScaleDN,Wu2018BlockDropDI}.
In conditional computing, subparts of the network are turned on or off dynamically based on the input image.   
This  allows to increase the network's capacity at training time without affecting the computational cost at inference. 
The same conditional behavior applies in early exiting but across the depth dimension: 
The prediction can be finalized early in the network for simple images, avoiding unnecessary further computing.

In particular, Mixture of Experts (MoE) have gained a lot of traction in recent years for conditional computing. 
For instance, transformers with a massive number of parameters are now becoming the new normal for natural language processing~\cite{Fedus2021Switch, Clark2021Scaling, Lepikhin2021Gshard, Riquelme2021ScalingVW}. 
Similar models are also starting to emerge in computer vision, leveraging extremely large datasets and numerous routing decisions~\cite{Riquelme2021ScalingVW}.
The success of these large-scale conditional models begs the question whether similar results are also achievable for datasets and architectures of a smaller scale, more commonly used by practitioners (e.g. ResNet-18 on ImageNet).
In this work, we introduce three ingredients to make simple single-gate MoE  competitive with other state-of-the-art MoE, across a variety of architectures  and dataset sizes. 
In particular, our training pipeline remains efficient and stable in all cases, in contrast to more complex dynamic routings~\cite{Zhang2021NeuralRB}, and avoids introducing new ad-hoc losses~\cite{zhang2022collaboration}. 
Specifically, we make the following contributions:
\begin{itemize}
\itemsep -0.1em 
    \item In \hyperref[sec:modelsum]{Section \ref{sec:modelsum}}, we introduce a single-gate MoE that consistently outperforms its non-mixture counterparts on various architectures. A key improvement in our proposed model is a \textit{base network} branch whose features facilitate the initial expert selection. We also show that this base model acts as an excellent regularizer when ensembled with specialized experts, improving the overall performance.
    \item  In \hyperref[alg:training]{Algorithm \ref{alg:training}}, we also formulate a \textit{simple and efficient} training procedure for the model, which is both stable and asynchronous.
    These results indicate that simple single-gate MoE is a promising direction to enable conditional computing for both small training- and inference- computational budgets.
    \item Finally, in \hyperref[sec:earlyexit]{ Section \ref{sec:earlyexit}},  we propose and evaluate a simple threshold rule to dynamically adapt the computational budget at inference, without retraining, which combines early-exiting through the base model and selecting a dynamic number of experts per sample.
\end{itemize}

\section{Proposed Model}\label{sec:methodology}
\subsection{The Mixture of Experts Setup for Image Classification}
\label{sec:setup}
A Mixture of Experts (MoE) consists of a set of $K$ \emph{experts}, $(e_k)_{1\dots K}$, each outputting a distribution over the  target classes; The execution of these experts is conditioned by the  \emph{gate} $\gate$ (or \emph{router}), which outputs a probability distribution over the set of experts. 
The total likelihood of the model on the training dataset $\mathcal D$, which we want to maximize, is expressed as:
\begin{align}
    \label{eq:mle}
    \mathcal L (D) = \mathop{\mathbb{E}}_{(x, y) \sim D} \left[ \sum_{k=1}^K g(k | x)\ e_k(y | x) \right]
\end{align}

A successful MoE relies on the gate learning a decomposition of the input space across $K$ clusters, such that experts specialize on the resulting subsets; The key underlying assumption is that this compositional approach outperforms a single model trained on the entire dataset.
At inference time, the gate is thresholded, such that only one - or few - experts are executed, to control the accuracy/efficiency trade-off. 
Unfortunately, the standard MoE suffers from three major issues: 
\textbf{(i)} Because the experts only have a local view of the training set, regulated by the gate, their mixture is more prone to overfitting than a single model trained on the whole dataset~\cite{Rosenbaum2019RoutingNA}.
\textbf{(ii)} Jointly training the gate and experts raises a chicken-and-egg problem: 
The gate has to route samples to the experts most likely to classify them successfully, but weaker experts need data to improve. 
This problem often leads to the gate collapsing, i.e., only feeding input samples to very few experts, which defeats the purpose of using an MoE in the first place. 
\textbf{(iii)} The initial data subsets defined by the gate strongly influences the expert training. Thus, a naive random initialization may even further worsen the gate collapse issue if, for instance, an expert is heavily favored over others at initialization.

In this work, we propose two key changes to the MoE framework to alleviate the aforementioned issues and improve performance of simple single-gate MoE models. 
\textbf{First}, we introduce a novel generic knowledge branch, which we refer to as the \emph{base model}: This module is trained on the whole dataset, and we use it \textbf{(i)} to tackle potential overfitting: It acts as a form of regularization for the experts by ensembling their outputs with this initial base prediction;  
\textbf{(ii)} to initialize the gate, using the feature space induced by the base model for clustering the training samples. 
and \textbf{(iii)} as an early-exiting branch that avoids executing any expert when not necessary, and is conditionally activated based on the input image.
\textbf{Second}, we describe a simple, lightweight training scheme that first initializes the experts' subsets by clustering the base model's embeddings and then keeps the gate and experts independent during training in order to avoid the gate collapse issue. 

\begin{figure}[ht]
    \centering
    \begin{minipage}[c]{0.35\textwidth}
    \centering
    \includegraphics[width=\textwidth]{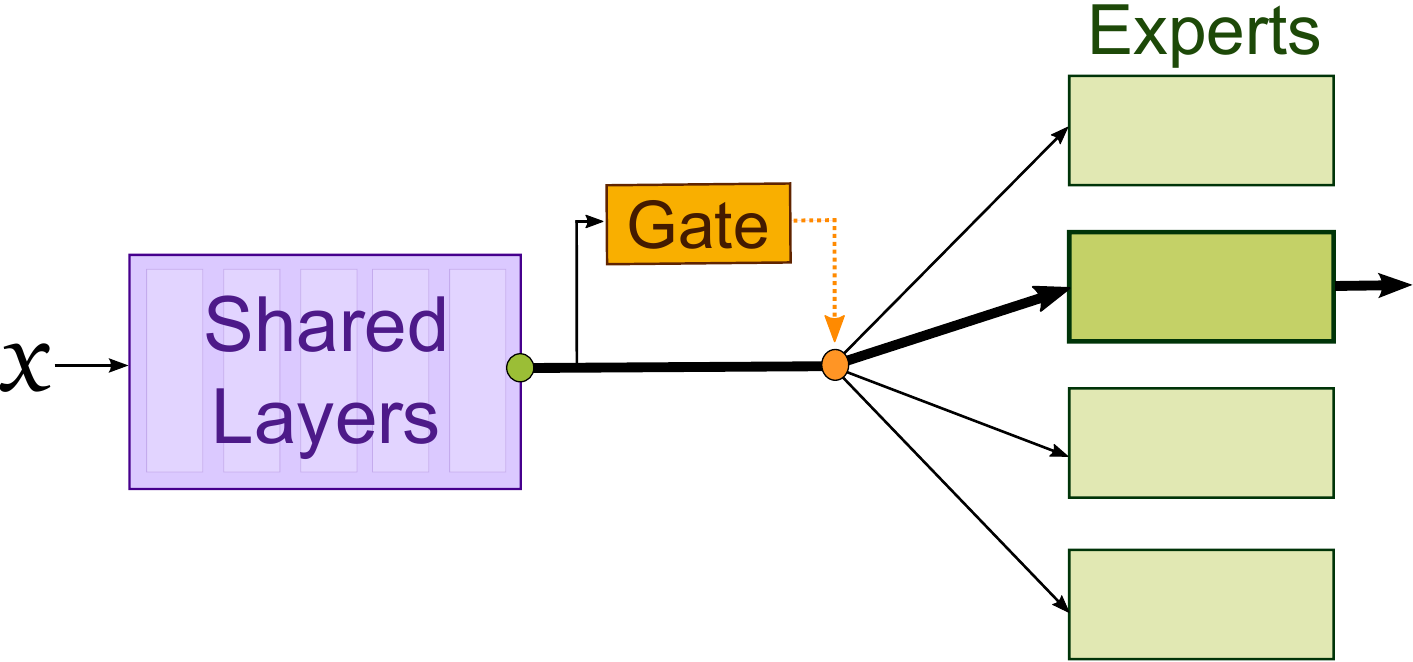}
    
    \textbf{(a)} Single-gate MoE, e.g. \cite{Mullapudi2018HydraNetsSD,Yan2015HDCNNHD,Gross2017HardMO,zhang2022collaboration}
    \end{minipage}
    \hspace{0.021\textwidth}
    \begin{minipage}[c]{0.61\textwidth}
    \centering
    \includegraphics[width=\textwidth]{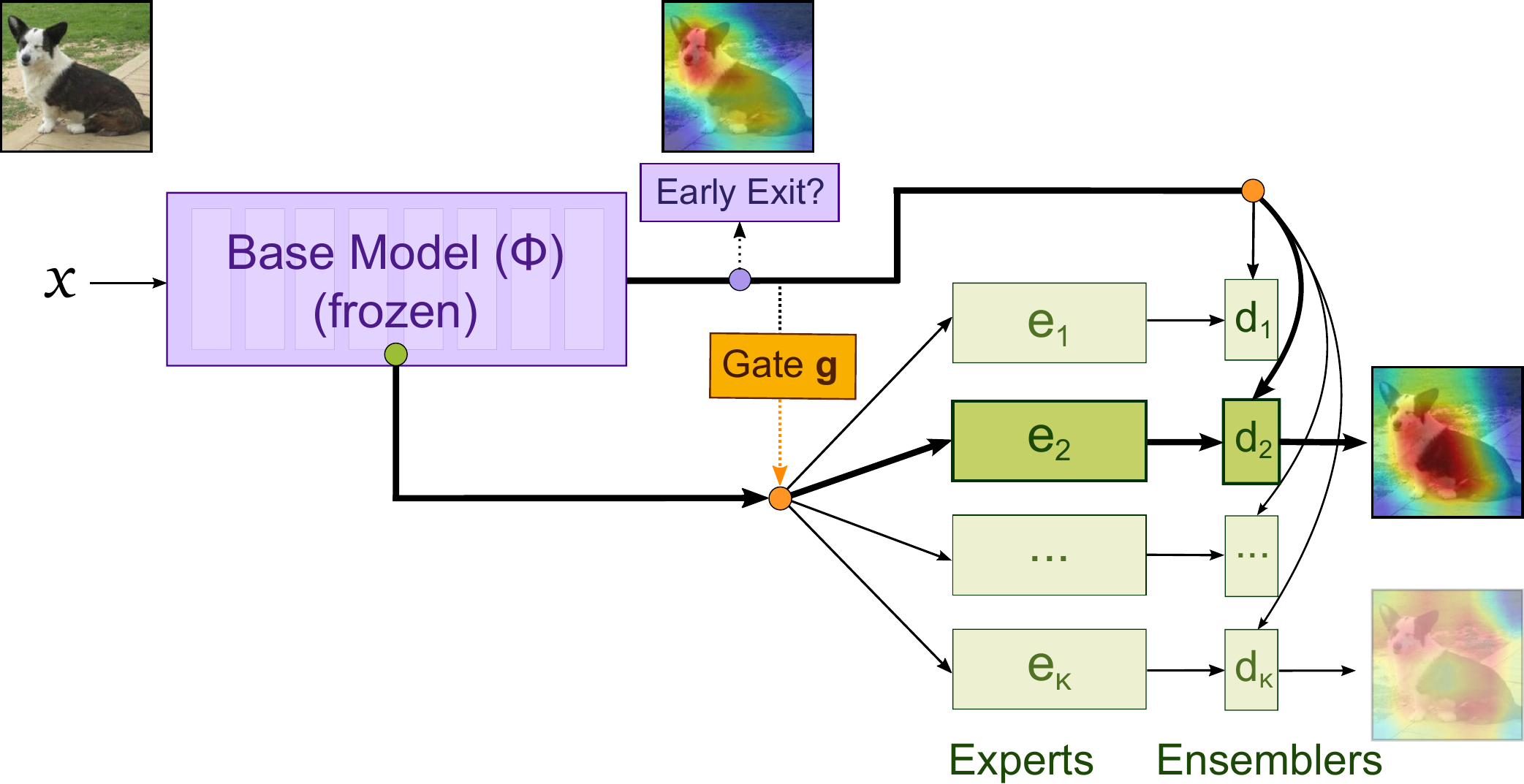}
    \textbf{(b)} Proposed model
    \end{minipage}
    \caption{\label{fig:model} \small (\textit{left}) MoE define a gate, $g$, that selects ({\footnotesize \textcolor{orange}{$\newmoon$}}) which expert to execute  based on the current representation  ({\footnotesize \textcolor{LimeGreen}{$\newmoon$}}) of  input $x$. 
    At inference, a unique expert is picked (in \textbf{bold}). 
    (\textit{right}) our proposed architecture maintains a full-depth base model, $\phi$, which is (i) ensembled with the expert output, (ii) used as inputs to the experts and gate, and (iii) acts as an early exit ({\footnotesize\textcolor{Thistle}{$\newmoon$}}) at inference.
    Grad-CAM~\cite{jacobgilpytorchcam} visuals reveal that the selected expert  focuses on fine-grained details,
    while the base model attends to general features. 
    The other non-selected experts produce poorly focused activation maps. 
    }
\end{figure}

In \hyperref[sec:modelsum]{Section \ref{sec:modelsum}}, we describe the proposed model's key components and training scheme; The model architecture is summarized in \hyperref[fig:model]{Figure \ref{fig:model}}. 
Then, in \hyperref[sec:earlyexit]{Section \ref{sec:earlyexit}} we describe a simple conditional computing mechanism to obtain even more computationally-efficient models. 

\subsection{Model Summary}

\label{sec:modelsum}

\vspace{-0.35cm}
\paragraph{Architecture.} The \textbf{base model} $\phi$ is a simple, ideally lightweight, network trained on the whole dataset, and is executed for every input. Its purpose is multi-fold: First, it is ensembled with the selected expert. While previous works~\cite{Yan2015HDCNNHD,Mullapudi2018HydraNetsSD} often ensemble specialized experts together in MoE, we show in our experiments that ensembling one expert with this non-specialist branch is consistently more beneficial.
Second, the base model acts as an early exit output at inference time, avoiding redundant expert computations for the easier samples (see \hyperref[sec:earlyexit]{Section \ref{sec:earlyexit}}). 
Finally, we reuse the early layers of the base model as inputs to the gate and the experts which allows us to reduce computational load even further.

The \textbf{gate} $g$ is a simple linear layer taking as input the pre-logits of the base model. At training time, it outputs a probability distribution over experts, $g(k | x)$, allowing for direct backpropagation through these weights. At inference, we only select and execute the most probable expert, i.e., $g_{\text{test}}(k | x) = \mathbbm{1}{( k = \arg\max_{k'} g(k' | x) )} $; $\mathbbm{1}(\cdot)$ being the indicator function. 
We also discuss in \hyperref[sec:earlyexit]{Section \ref{sec:earlyexit}} how we can dynamically select the number of active experts, rather than always defaulting to the top-1 expert.

\textbf{Experts } are neural networks whose input is an intermediate feature map of the base model. 
This design choice yields two benefits: \textbf{(i)} The experts' early features are shared and frozen, which reduces the number of trainable parameters and reduces the risk of experts overfitting (in particular when training on small datasets); and \textbf{(ii)} this allows the model to reuse computations from the base model at inference time, further improving efficiency.

Finally, \textbf{ensemblers} are shallow neural networks, one for each expert, combining outputs of the base model and the expert selected by the gate. We experiment with both stacking and bagging ensembling methods. In the text, we also refer to $e'_k(y | x) = d_k(\phi(y | x); e_k(y | x))$ as the classification output of the ensembler $d_k$, which ensembles the $k$-th expert and base model $\phi$.

\vspace{-0.3cm}
\paragraph{Training Procedure.}
We summarize our asynchronous training scheme in \hyperref[alg:training]{Algorithm \ref{alg:training}}, in which the gate and experts are trained independently in parallel. 
This training procedure relies on three key insights: 
 \textbf{First}, to avoid gate collapse, we keep the gate weights fixed while training the experts. This makes the model heavily dependent on the gate initialization. Thus, to define a meaningful initial gate $g_0$, we cluster the pretrained base model's embeddings using  $K$-means~\cite{MacQueen1967SomeMF}. 
A similar initialization scheme has been used in the hierarchical classification literature~\cite{Gross2017HardMO,Mullapudi2018HydraNetsSD}; In contrast, we do not restrict this initial clustering step to be a hard assignment to a unique expert, nor to be on a per-class basis. 
 
A \textbf{second} issue stems from uncalibrated outputs~\cite{Guo2017OnCO,Minderer2021RevisitingTC}: Training an ensembler $d_k$ jointly with its expert $e_k$ often leads to $d_k$ heavily favoring the base model, preventing the expert from specializing. This behavior is likely due to the base model being overly confident on many training samples: 
In fact, this is particularly apparent on small datasets where the  base model is already close to perfectly fitting the training set, e.g., on CIFAR-100. 
To avoid this problem, we only start training $d_k$ \emph{after} fully training the corresponding expert $e_k$.

\textbf{Thirdly}, because the experts are initialized with the base model's pretrained weights, but are then trained on a specialized subset of the data given by the gate, they might ``forget'' classes they never see. This is similar to \emph{catastrophic forgetting}~\cite{Kirkpatrick2017OvercomingCF,Ramasesh2021AnatomyOC}. 
While the proposed ensemblers partially alleviate this issue by providing additional regularization, we find that it is often beneficial to also route non-assigned samples the experts: 
Specifically, in step 4 and 5 of \hyperref[alg:training]{Algorithm \ref{alg:training}}, the gate $g_0$ is "smoothed" using the transformation: $\Gamma: \cdot \mapsto \text{clip}(\cdot, \gamma, 1.)$, where $\gamma$ is a hyperparameter. We experimented with \textbf{(i)} using the smoothed gate weights to re-weight the loss of all samples, including negative ones (as portrayed in \textbf{Algorithm 1}) or \textbf{(ii)} using these gate weights as sampling probabilities when forming the training batch. Both yield similar results, and while (i) is simpler to implement, we find that (ii) is more practical for large datasets, as it often leads to faster convergence, hence reduced training times.

\begin{algorithm}[ht]
\begin{algorithmic}
\Require Training dataset $\mathcal D$, base model $\phi$, gate $g$, experts $e_{1\dots K}$, ensemblers $d_{1\dots K}$
\State 1. Train the \textbf{base model} $\phi$ on dataset $\mathcal D$ (or use an off-the-shelf pretrained model)
\State 2. Cluster the base model embeddings using $K$-means, obtaining centroids $c_{1 \dots K}$
\State Define the \textbf{initial gate} $g_0 : (x, k) \mapsto m(\phi(x), c_k)$ where $m$ is a similarity function and is   such that the weights across clusters sum to 1 for a given sample $x$. 
\State 3. Train the \textbf{gate} $g$ by minimizing the KL divergence $\text{KL}(g_0, g)$
\For{$k = 1 \text{ to } K$ (in parallel)}
\State Initialize the $k$-th expert from the base model's weights: $\theta_{e_k} \leftarrow \theta_{\phi}$
\State 4. Train \textbf{expert} $e_k$ following \eqref{eq:mle}, i.e., by maximizing $\sum_{(x,y)\sim D} g_0(k | x)\ e_k ( y | x)$
    \State  5. Train \textbf{ensembler} $d_k$ by maximizing $\sum_{(x,y)\sim D} g_0(k | x)\ d_k ( \phi(y | x); e_k(y | x))$
\EndFor
\end{algorithmic}
\caption{\label{alg:training}Proposed asynchronous training scheme}
\end{algorithm}

\paragraph{An alternative joint training scheme.}
We also consider extending the training scheme to handle joint end-to-end training of the gate and experts by using the Expectation Maximization (\textbf{EM}) algorithm~\cite{Dempster1977MaximumLF} to alleviate the "chicken-and-egg" problem. EM alternates between two steps: \textbf{(E)} computing new gate weights, updated based on the current experts' performance, and \textbf{(M)} separately training the experts according to this new assignment, while forcing the gate to match it.
Taking into account training costs, it is beneficial to keep a low number of \textbf{E} steps ($N_E$) as every update of the posterior requires synchronization across all experts. In fact, one can show that Algorithm 1 is equivalent to setting $N_E = 0$.
In practice, we observe that larger values $N_E$ can lead to higher accuracies (e.g., on tiny-ImageNet: \textbf{+1.37\%} without ensemblers, and \textbf{+0.43\%} with ensemblers). However, the improved performance is often not worth the higher training costs, hence we only report results using  \textbf{Algorithm 1} in our experiments section. 
We describe the derivation of the EM variant and experiments in more details in the \textbf{supplemental material}.

\subsection{Anytime Inference via Early-Exiting and Dynamic Ensembling}
\label{sec:earlyexit}
By design, our framework integrates a straightforward option for early-exiting by directly outputting the base model's predictions in easy cases, to improve computational efficiency further. 
Following previous early-exiting literature~\cite{Teerapittayanon2016branchynet,Xin2020DeeBERTDE,Huang2018MultiScaleDN}, our model decides whether to early-exit or to execute the gate-selected expert by thresholding the base model's confidence at inference time. We also consider other early-exit designs in the \textbf{supplemental material}.
Additionally,  it is clear from \hyperref[eq:mle]{Equation \ref{eq:mle}} that MoE can be viewed as an ensemble of experts weighted by the gate, rather than using only the top-1 expert as is usually done for efficiency purposes. 
Similar to ~\cite{Yan2015HDCNNHD}, we  propose to threshold the gate outputs to determine which experts to include dynamically at inference. 
In order to combine both the early-exiting and expert ensembling behaviors under a unique  thresholding rule, we introduce the quantity  $\mathbf{\alpha_{k}(x) = g(k | x) (1 - \max_y\phi(y | x))}$. From a probabilistic perspective, $\alpha_k(x)$ can be  interpreted as the joint probability that the sample $x$ is not early-exited, \emph{and} that the gate routes $x$ to the $k$-th expert.
Intuitively, if this quantity is below a certain threshold for all experts, it means that the base model has a high confidence and the gate does not confidently route the sample to any expert; thus we should early exit.
Thus, given a trained gate $g$ and experts (with their ensemblers) $e'_k$,  we define the \textbf{anytime model} $p^{\text{at-}\tau}$, which combines both early-exiting and dynamic experts ensembling, as follows: 
\begin{align}
ee(x) &= 1 \text{ iff } \forall k \in [1, K],\ \alpha_{k}(x)  < \tau \label{eq:ee} \\ 
p^{\text{at-}\tau}(y|x) &= ee(x) \phi(y  |x) + (1 - ee(x)) \sum_{k=1}^K \ \mathbbm{1}{\left( \alpha_{k}(x) \geq\tau \right)} \ g(k | x) \ e'_k(y | x) \label{eq:ee-prob}
\end{align}
where $\mathbbm{1}{(\cdot)}$  is the indicator function, and $\tau \in [0, 1]$ is a hyperparameter.
We show in experiments that varying  $\tau$ allows the model to quickly achieve a wide range of computational budgets at inference time, without any retraining.  

\section{Related Work}\label{sec:related_work}
\textbf{Conditional computing} aims to learn a sparse connectivity pattern conditioned on the input sample. 
To achieve such pattern, many works add a mixture of expert layers at several stages throughout the network~\cite{Eigen2014LearningFR,Wang2019DeepMO,Wang2018SkipNetLD,Cai2021DynamicRN,Rao2019RuntimeNR,Zhang2021NeuralRB,McGill2017DecidingHT}. 
 Unlike single-gated MoE, the increased number of routing decisions incurs some practical drawbacks: (i) At inference, a new submodel has to be loaded in memory for every routing decision, which becomes increasingly cumbersome as the number of gates increases and (ii) all gates and experts have to be trained synchronously which leads to huge models to train and training instabilities. 
This often leads to complex training pipelines, e.g. relying on reinforcement learning~\cite{Wang2018SkipNetLD,Cai2021DynamicRN,Rao2019RuntimeNR} to learn the routing mechanism.
More recently,~\cite{Zhang2021NeuralRB} has proposed a simpler k-means-like routing mechanism that evolves during training via moving average. However, they also report that the training pipeline requires large batch sizes, and is prone to mode collapse.

Simpler \textbf{single-gate mixture of experts} have also been successfully applied to  neural networks for various applications such as image classification~\cite{Ahmed2018BranchConnectIC}, detection~\cite{Lee2020MERM} , retrieval~\cite{Gross2017HardMO} and scene recognition~\cite{Kim2018HierarchyOA}. 
More recently,~\cite{zhang2022collaboration} has shown that such models  can achieve significant accuracy/efficiency gains in the large-scale regime. However, their training pipeline relies on several unclear heuristics. 
Orthogonal to these, hierarchical classification is a subclass of MoE, in which the routing is learned on a \emph{per-class} basis, aiming to route all samples of a ground-truth class to the same expert.
Several works~\cite{Marszalek2008ConstructingCH,Goo2016TaxonomyRegularizedSD,Bertinetto2020MakingBM,Puigcerver2021ScalableTL} directly leverage an external class taxonomy (such as WordNet~\cite{Miller1990IntroductionTW}). A follow-up line of thought extracts such information from a pretrained classifier~\cite{Yan2015HDCNNHD,Mullapudi2018HydraNetsSD}, or even learns the optimal taxonomy jointly with the image representations~\cite{Ahmed2016NetworkOE,Li2021MMFMM}.
Such models have been shown to improve the efficiency/accuracy trade-off in classification tasks. 
However, this class-based routing is a limiting assumption, and per-sample routing has been shown to outperform hierarchical classification models when correctly parametrized~\cite{Ahmed2018BranchConnectIC,Kim2018HierarchyOA}.

Finally, MoE can be seen as an \textbf{ensembling} technique whose weights are learned by the gate. While it is common to assume each sample is routed to a unique expert to maximize efficiency, some  works~\cite{Yan2015HDCNNHD,Mullapudi2018HydraNetsSD,Gross2017HardMO} have considered combining several experts to boost accuracy. 
In contrast, we show that combining one specialized expert with the generic knowledge base model with simple ensemble methods such as averaging or linear stacking~\cite{Sill2009FeatureWeightedLS} is generally  more efficient than ensembling multiple specialized experts.

\section{Experiments}\label{sec:experiments}
We perform experiments on datasets of different scales: CIFAR-100~\cite{Krizhevsky2009LearningML},  tiny-ImageNet~\cite{tinyimagenet} (a downscaled subset of ImageNet with 200 classes and 110k images), and ILSVRC2012~\cite{Russakovsky2015ImageNetLS}.  

We use ResNets~\cite{He2016DeepRL} with different depths as our main backbone architecture. 
For CIFAR-100 and tiny-ImageNet, we use a modified variant of ResNets which eliminates the first two downscaling operations (strided convolution and max-pooling), commonly used in the literature~\cite{Uddin2021SaliencyMixAS,Kim2020PuzzleME,Zhang2021NeuralRB};  We dub it ``\emph{tiny-ResNet}'' or \textbf{tr} for short.
We follow previously established training pipelines to train our baseline models, specifically \cite{Devries2017ImprovedRO} for CIFAR-100 and \cite{Li2021BoostingDV} for tiny-ImageNet. 

For ImageNet, we additionally perform experiments on MobileNetv3-small~\cite{Howard2019SearchingFM}, and use the standard checkpoints provided in torchvision's model zoo~\cite{modelzoo} as base models.
In all of our experiments, we initialize the experts with pretrained weights from the base model and train them with the same hyperparameters and data augmentations as the baseline, although using fewer training iterations (200 epochs for CIFAR100, 100 for tiny-ImageNet and 40 for ImageNet). The features of the base model are kept frozen.

In this section, to assess the benefits of our proposed model and training scheme,  we compare our proposed method to\textbf{ (i) }the backbone models at different depths,\textbf{ (ii)} an ensembling baseline with equivalent computational cost, \textbf{(iii) }hierarchical classification, and \textbf{(iv)} two recent dynamic routing works~\cite{Wang2019DeepMO,Zhang2021NeuralRB}. We also report results of an  ablation experiment on using different ensembling methods.
Finally, we report detailed hyperparameters used for the experiments, and further ablations in the supplemental material.

\subsection{Results on Small and Medium-scale Datasets}
We first report results on CIFAR-100 and tiny-ImageNet for tiny-ResNets of different depths in  \hyperref[table:cifar]{Table \ref{table:cifar}}. 
All results are reported with 20 experts, branching off the base model after the third residual block, and \textit{without any early-exiting or dynamic ensembling}. 
These results show that even simple single-gate MoE can significantly improve the efficiency/accuracy trade-off over standard CNNs:  
Our proposed method consistently outperforms the backbone network for an equivalent MAC count. 
The only downside is that MoE generally has a higher parameter count at training time. Nevertheless, our asynchronous training scheme allows us to train experts independently across multiple devices efficiently.
We also observe that using a deeper base model is often more beneficial than using deeper experts in terms of accuracy. 

\begin{table}[t]
    \centering
\resizebox{0.49\textwidth}{!}{%
\begin{tabular}{|c|c||c||c||c|c|}
   \hline
  Base  & \multirow{2}{*}{Expert} & \multirow{2}{*}{top-1 acc} &  MACs & \multicolumn{2}{c|}{\#params x 1e7}\\
  \cline{5-6}
  Model &   &  & (x 1e9) & inference & trainable\\
  \hline
  \footnotesize tr18 baseline & - & 77.95 & 0.56 & 1.12 & 1.12 \\
  \hline
  \hline
  \multirow{2}{*}{tr10}  &  tr10 & 77.96 ± 0.20 & 0.37 & 0.96 & 9.29 \\
  \cline{2-6}
   &  tr18 & 78.78 ± 0.22 & 0.52 & 1.55 & 21.1 \\
   \hline
  \hline
  \footnotesize tr34 baseline &  - & 78.60 & 1.16 & 2.13 &  2.13\\
  \hline
  \multirow{2}{*}{tr18}  &  tr10 & 79.78 ± 0.05 & 0.67 & 1.58 & 9.29 \\
  \cline{2-6}
    &  tr18 & 79.90 ± 0.22 & 0.82 & 2.17 & 21.1 \\
  \hline
  \hline
  \footnotesize tr50 baseline &  - & 80.10 & 1.30 & 2.37 &  2.37 \\
  \hline
  tr34  &  tr10 & 80.48 ± 0.17 & 1.28 & 2.59 & 9.29 \\
  \hline
\end{tabular}
}
~
\resizebox{0.49\textwidth}{!}{%
\begin{tabular}{|c|c|c|c|c|c|c|}
  \hline
  Base  & \multirow{2}{*}{Expert} & \multirow{2}{*}{top-1 acc} &  MACs & \multicolumn{2}{c|}{\#params x 1e7}\\
  \cline{5-6}
  Model &   &  & (x 1e9) & inference & trainable\\
  \hline
  \footnotesize tr18 baseline & - & 60.42 & 2.22 & 1.13 &  1.13 \\
  \hline
  \multirow{2}{*}{tr10}  &  tr10 & 60.58 ± 0.04 & 1.48 & 0.96 & 9.39 \\
  \cline{2-6}
    &  tr18 & 63.38 ± 0.05 &  2.09 & 1.55 & 21.2 \\
  \hline
  \hline
  \footnotesize tr34 baseline &  - & 63.39 & 4.64 & 2.14 & 2.14 \\
  \hline
  \multirow{2}{*}{tr18}  & tr10 & 64.46 ± 0.05 & 2.69 & 1.59 & 9.39 \\
  \cline{2-6}
    &  tr18 & 66.26 ± 0.05 & 3.30 & 2.18 & 21.2 \\
  \hline
  \hline
  \footnotesize tr50 baseline &  - & 63.24 & 5.19 & 2.39 &  2.39 \\
  \hline
  tr34  &  tr10 & 66.42 ± 0.15 & 5.11 & 2.60 & 9.39 \\
  \hline
\end{tabular}
}
    \caption{\label{table:cifar}Main CIFAR-100 (\textit{left}) and tiny-ImageNet (\textit{right}) results. Each number is reported over three random seeds. All settings have 20 experts, whose first three blocks are the (frozen) layers of the base model. We report accuracy and efficiency metrics (number of operations and number of parameters) across various base and experts architectures.}
\end{table}

\begin{figure}[t]
\centering 
\includegraphics[width=0.49\textwidth]{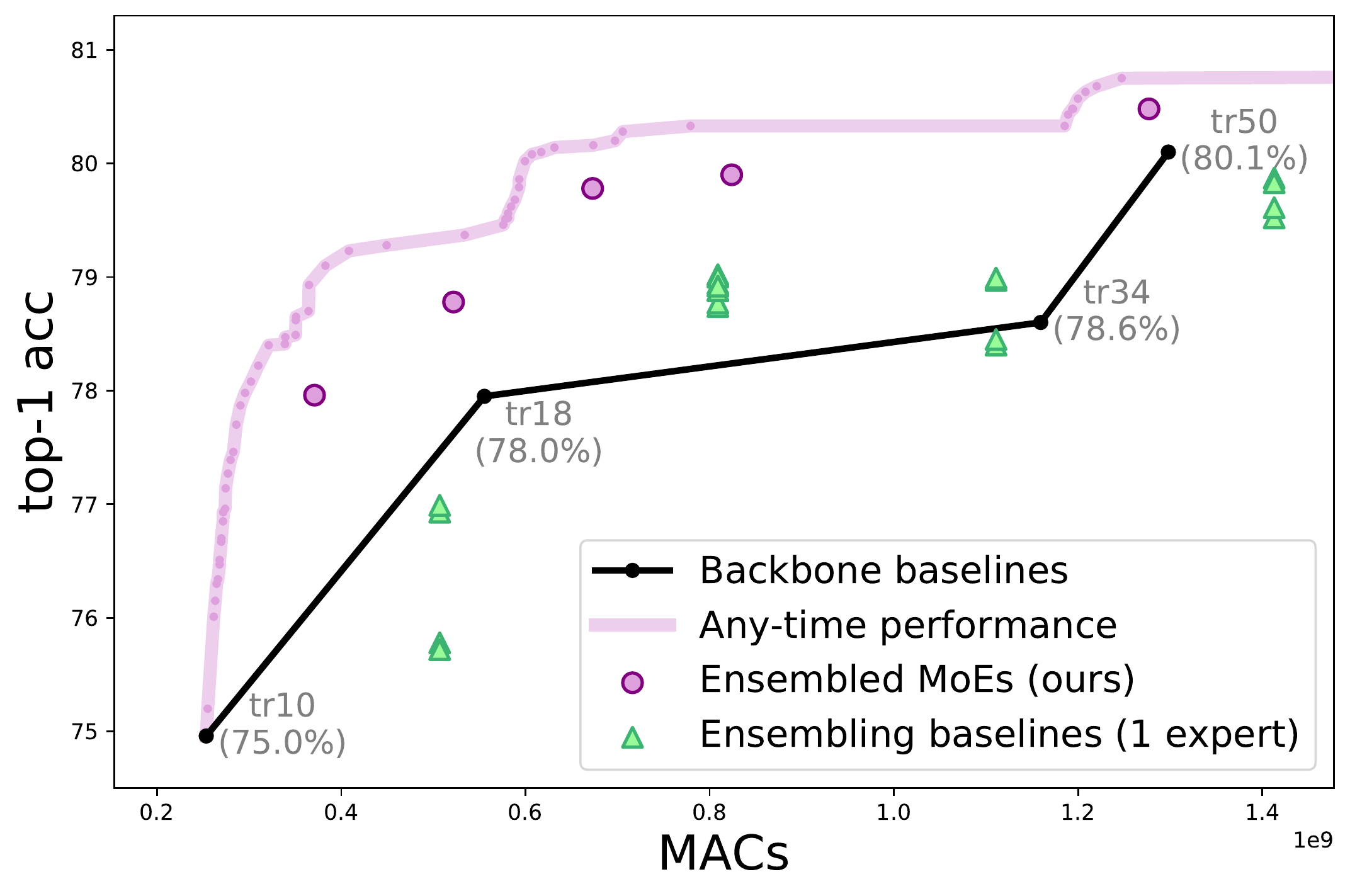}~\includegraphics[width=0.49\textwidth]{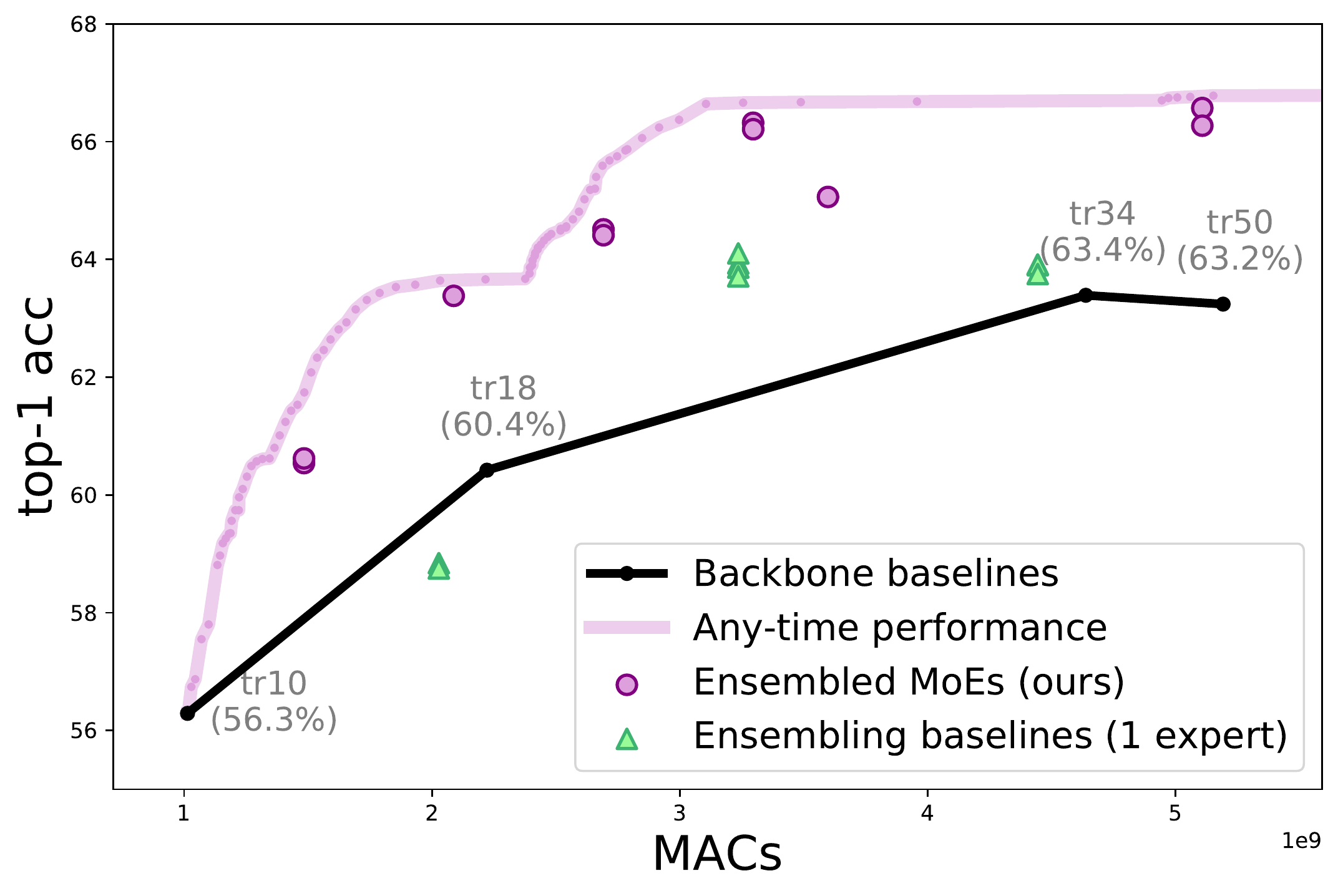}
\caption{\label{fig:gatethreshold}Accuracy vs MACs performance of our ``anytime'' variant models using a simple thresholding rule on CIFAR-100 (\emph{left}) and tiny-ImageNet (\emph{right}). }
\end{figure}

\begin{wraptable}[11]{r}{0.48\textwidth}
    \centering
\resizebox{0.47\textwidth}{!}{%
\begin{tabular}{|c||c||c|}
   \hline
  tr18-tr18  & CIFAR100 & tiny-ImageNet \\
  \hline
  \hline
  base model  & 77.95  & 60.42  \\
  \hline
  \hline
  baseline (1 expert)  & 78.99 ± 0.29  & 63.83 ± 0.08  \\
  \hline
  5 experts  &  79.67 ± 0.14 &  65.42 ± 0.15  \\
  10 experts  & 79.84 ± 0.13& 65.72 ± 0.17 \\
  20 experts  & \textbf{79.90} ± 0.22 &  \textbf{66.26} ± 0.05 \\
  \hline
\end{tabular}
}
    \caption{\label{table:numexperts}Impact of the number of experts on the model accuracy and comparison to the ensembling baseline (equivalent to using only one expert in our model). }
\end{wraptable}

\vspace{-7pt}
\paragraph{Comparison to ensembling.} A natural baseline to compare to is to ensemble the base model with a unique ``expert''  trained on the whole dataset:  The resulting model has the same cost as its MoE counterpart, minus the negligible cost of the linear layer gate. 
We also analyze the impact of the number of experts on the model: Adding more experts leads to higher specialization, but also more potential routing errors; thus, it is not evident that increasing the number of experts benefits the model.
We report the corresponding results in \hyperref[table:numexperts]{Table \ref{table:numexperts}}: 
All MoE outperform the ensembling baseline, which shows the benefit of specialized experts. Hower, the impact of going from 10 to 20 experts is minimal: The benefit of splitting the data into specialized subsets starts to  fade above 10 experts for these datasets. 

\paragraph{Anytime Inference Models.}
\label{sec:earlyexitresults}
We explore the effect of the anytime inference model introduced in \hyperref[sec:earlyexit]{Section \ref{sec:earlyexit}}. 
To fully understand the scope of this dynamic behavior, we evaluate the accuracy of all  models across various thresholds. We then plot the convex envelope of this set of curves, as shown in \hyperref[fig:gatethreshold]{Figure \ref{fig:gatethreshold}}.  
We observe that dynamically deciding for each sample whether to  early exit through the base model or to use one or more experts consistently improves the overall accuracy/efficiency trade-off. 
Furthermore, this simple thresholding rule allows us to quickly adapt the model's computational budget without retraining.

\subsection{Results on ImageNet}

\paragraph{Main results.}
We report results on ImageNet experiments for ResNet and MobileNet backbones in \hyperref[table:imagenet_resnet]{Table \ref{table:imagenet_resnet}}.  %
Our previous observations still hold: The MoE model outperforms both the ensembling baseline and the backbone, and early-exiting based on the base model's confidences helps further reduce computations for a limited drop in accuracy.

We also note that, unlike ResNet18, ResNet34's MobileNetv3's performance start to saturate and even slightly decreases with more experts. This behavior implies that the optimal number of experts is not only a property of the dataset but also of the architecture of both the experts and the base model.

\begin{table}[h]
\begin{minipage}[t]{0.45\textwidth}
    \centering
\resizebox{\textwidth}{!}{%
\begin{tabular}{|c||c|c|c|}
   \hline
  ResNet18& None & $\tau = 0.75$ & $\tau=0.5$ \\
  \hline
  \hline
  baseline (1 expert)  & 71.50 & 71.50 & 71.13 \\
  \hline
  4 experts  & 72.17 & 72.11 & 71.68\\
  \hline
  20 experts & \textbf{72.38} &\textbf{ 72.38} & \textbf{71.73} \\
  \hline
  MACs & 2.64e9 &  2.18e9 & 2.03e9 \\
  \hline
\end{tabular}
}

\centering \small \textbf{(a)} ResNet18 base model (\textbf{69.76}\% accuracy, 1.82 GMACs) and experts
\end{minipage}
\begin{minipage}[t]{0.26\textwidth}
    \centering
\resizebox{\textwidth}{!}{%
\begin{tabular}{|c|c|c|}
   \hline
   None & $\tau = 0.75$ & $\tau=0.5$ \\
  \hline
  \hline
  74.33 & 74.29 & 74.00 \\
  \hline
   75.03 & 74.89 & 74.47 \\
  \hline
   \textbf{75.05} & \textbf{74.92} &\textbf{ 74.56} \\
  \hline
   5.64e9 &  4.42e9 &  4.06e9 \\
  \hline
\end{tabular}
}

\centering \small \textbf{(b)}  ResNet34 base model (\textbf{73.31}\%, 3.66 GMACs) and experts
\end{minipage}
\begin{minipage}[t]{0.26\textwidth}
    \centering
\resizebox{\textwidth}{!}{%
\begin{tabular}{|c|c|c|}
   \hline
  None & $\tau = 0.75$ & $\tau=0.5$  \\
  \hline
  \hline
   68.06 & 68.13 & 68.15  \\
  \hline
   \textbf{68.60} & \textbf{68.59} & 68.44\\
  \hline
    68.58 & 68.53 & \textbf{68.46}\\
  \hline
   8.13e7 &  6.83e7 & 6.36e7 \\
  \hline
\end{tabular}
}

\centering \small \textbf{(b)} MobileNetv3-small base (\textbf{67.67}\%, 5.65e7 MACs) and experts
\end{minipage}

    \caption{\label{table:imagenet_resnet}Main results on ImageNet. Experts share 3 layers in the ResNet experiments, and 8 in the MobileNet ones (half of the model in both case). 
    We report our main results in the first column, without any early-exiting or dynamic ensembling. In the following columns, we report additional results with early-exiting obtained with different values of the  threshold $\tau$ on the base model confidence.}
\end{table}

\textbf{Comparison to Dynamic Routing baselines.}
In this section, we compare our model to two recent dynamic routing works.
In \hyperref[table:routing]{Table \ref{table:routing}} (\emph{left}) we compare to DeepMoE~\cite{Wang2019DeepMO} on Imagenet: The model is a  two-times wider ResNet-18 trained end-to-end with a sparsity constraint forcing the  router in each layer to only activate roughly half of the channels for each sample.
In \hyperref[table:routing]{Table \ref{table:routing}} (\emph{right}), we  compare to RMN~\cite{Zhang2021NeuralRB} on tiny-ImageNet: Each residual block is replicated 8 times (a total of $8^4$ different computational paths), and for each, a routing based on a moving average of initial k-means centroids is learned. 
We could not compare directly to RMN on ImageNet as \cite{Zhang2021NeuralRB}  uses a modified ResNet architecture including additional Squeeze\&Excite~\cite{Hu2020SqueezeandExcitationN} layers.
Nevertheless, in terms of relative accuracy improvement and MACs, we observe that our model yields comparable trade-offs, despite using a single gate, hence significantly fewer computational paths.
Furthermore, in contrast to both approaches, we can easily reach various   computational budgets without retraining via early-exiting.

\begin{table}[h]
    \centering
    
\resizebox{0.458\textwidth}{!}{%
\begin{tabular}{|c||c|c|c|c|}
   \hline
   ImageNet & \multirow{2}{*}{gates} & \multirow{2}{*}{acc.} & MACs & \#params\\
   Resnet-18 & &  & x1e9& (train) x1e9\\
   \hline
   ours &  1 & 72.17 & 2.64 & 5.10 \\
   \hline
   +early-exit $\tau=0.75$ & 1 & 72.11 & 2.18 & 5.10\\
   \hline
   \hline
    DeepMoE~\cite{Wang2019DeepMO} & 17 & 70.95 & 1.81 & 7.02 \\
   \hline
\end{tabular}
}
\resizebox{0.527\textwidth}{!}{%
\begin{tabular}{|c||c|c|c|c|c|}
   \hline
   \scriptsize tiny-ImageNet & \multirow{2}{*}{gates} & base & \multirow{2}{*}{acc.} & \multirow{2}{*}{MACs} & \#params\\
   \texttt{tr}18 & & model &   && (train)\\
   \hline
   ours (10 exp) &  1 & 60.42 & 64.66 & 2.69e9 & 5.98e7 \\
   \hline
   + early-exit ($\tau=0.75$) &  1 & 60.42 & 64.58 & 2.44e9 & 5.98e7 \\
   \hline
   RMN (no SE)~\cite{Zhang2021NeuralRB} & 5 & 61.78 & 64.30 & 2.22e9 & 9.02e7  \\
   \hline
\end{tabular}
}
    \caption{\label{table:routing} Comparison to recent dynamic routing literature. On the left, we compare our 4-experts ImageNet model with Wide-DeepMoE-18 from \cite{Wang2019DeepMO}. On the right, we compare our \texttt{tr18-tr10} 10 experts model with the tiny-ResNet18-based model of ~\cite{Zhang2021NeuralRB}, excluding additional Squeeze\&Excite layers. For \cite{Zhang2021NeuralRB}, we also report the corresponding base model accuracy as we could not reproduce their baseline training results to use in our experiments.}
\end{table}

\vspace{-0.5cm}
\subsection{Ablation experiments}
In the supplemental material, we report further ablation experiments on (i) different early-exiting procedures an (ii) an alternative joint training based on Expectation-Maximization.

\vspace{0.1cm}
\begin{wraptable}[11]{r}{0.54\textwidth}
\centering 
\resizebox{0.53\textwidth}{!}{%
\begin{tabular}{|c|c|c|}
\hline
top-1 acc & w/o ensembling & w/ ensembling \\
\hline
per-sample (\textbf{ours}) & \textbf{63.11} ± 0.11 & \textbf{65.72} ± 0.10 \\
\hline
per-class & 62.48 ± 0.13 & 63.85 ± 0.08 \\
\hline
\hline
 per-class + \emph{oracle}  & 68.8 ± 0.10 & 67.99 ± 0.04 \\
 \hline
\end{tabular}
}
\caption{\label{tab:hierarchy}Comparison to hierarchical classification. We introduce a per-class variant of our model following \cite{Mullapudi2018HydraNetsSD,Yan2015HDCNNHD}, and an oracle variant in which the gate follows the true class-to-expert distribution. Results are reported on tiny-ImageNet with 10 experts, in the \texttt{tr}18-\texttt{tr}18 configuration.}
\end{wraptable}

\textbf{Per-sample vs per-class clustering initialization.} Hierarchical classifiers  are single-gated models whose routing is learned on a strict \emph{per-class} basis. In the literature, the class taxonomy is either based on external knowledge~\cite{Marszalek2008ConstructingCH,Goo2016TaxonomyRegularizedSD,Bertinetto2020MakingBM,Puigcerver2021ScalableTL}, clustering pretrained embeddings~\cite{Mullapudi2018HydraNetsSD,Yan2015HDCNNHD}, or  jointly learned alongside the features~\cite{Ahmed2016NetworkOE}.
We implement a per-class variant of our model, following the clustering process of  \cite{Mullapudi2018HydraNetsSD,Yan2015HDCNNHD}.

In \hyperref[tab:hierarchy]{Table \ref{tab:hierarchy}}, we show that per-sample routing always outperforms its per-class counterpart. 
Furthermore, we observe a significant accuracy gain if we re-evaluate the per-class model using the samples' ground-truth classes as an oracle for predicting the correct expert. This indicates that learning the per-class routing is the main bottleneck. 
In fact, a sample $(x, y)$ incorrectly mapped to an expert that has never seen class $y$ is very detrimental to accuracy, even if this is partially compensated by the effect of ensembling, 
In contrast, per-sample routing allows several experts to gain knowledge about the same class and introduces a more flexible notion of diversity among experts.

\paragraph{Analysing the Ensemblers' performance}

\begin{wraptable}[9]{r}{0.5\textwidth}
\vspace{-0.2cm}
    \centering
\resizebox{0.47\textwidth}{!}{%
\begin{tabular}{|c||c|c|c|}
   \hline
  base - expert  & tr10-tr10 & tr10-tr18 & tr18-tr18 \\
  \hline
  \hline
  base model  & 56.29 & 56.29 & 60.42 \\
  \hline
  \hline
  no ensembling  & 57.80 & 59.53 & 63.82 \\
  \hline
  top-2 experts &  58.69 & 60.55 & 64.19 \\
  stacking  & \textbf{60.84} &  64.63 &  66.29 \\
  bagging  & 60.54 & \textbf{64.77} & \textbf{66.32} \\
  \hline
\end{tabular}
}
    \caption{\label{table:decoder}Impact of the ensembler design on accuracy (tiny-ImageNet, 20 experts) }
\end{wraptable}

In \hyperref[table:decoder]{Table \ref{table:decoder}}, we compare results for different ensembling scheme with equivalent computation costs:   
Even without ensembling, MoE outperforms the base model. Nevertheless, all ensembling methods perform strictly better than trusting the selected expert only. 
Second, we observe that \textbf{stacking} and \textbf{bagging} always outperform ensembling the top-2 experts, which shows the benefit of ensembling with the base model rather than another specialized expert. 
We use \textbf{bagging} in our experiments as it is simpler, non-parametric, and does not require training.

\paragraph{Qualitative Analysis of The Experts' Specialization Pattern.}

\begin{wrapfigure}{r}{0.45\textwidth}
\vspace{-0.1cm}
\setlength{\fboxsep}{0pt}%
\setlength{\fboxrule}{1pt}%
\centering
\includegraphics[trim={0, 0.2cm, 2.5cm, 4.4cm}, clip, width=0.18\textwidth, height=0.11\textwidth]{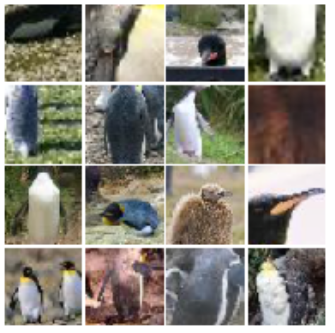}
~
\includegraphics[trim={0, 0, 2.5cm, 4.8cm}, clip, width=0.18\textwidth, height=0.11\textwidth]{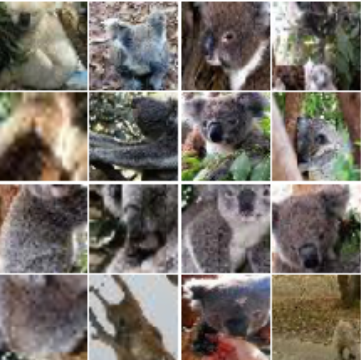}

\small \textbf{(a) }The class \texttt{king-penguin} (left) co-occurs with other animals (right) for full-view images\dots

\includegraphics[trim={0, 0.35cm, 2.5cm, 4.4cm}, clip, width=0.18\textwidth, height=0.11\textwidth]{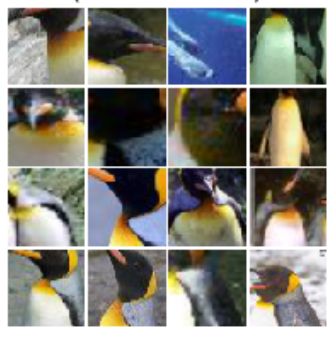}
~
\includegraphics[trim={0, 0, 2.5cm, 4.5cm}, clip, width=0.18\textwidth, height=0.11\textwidth]{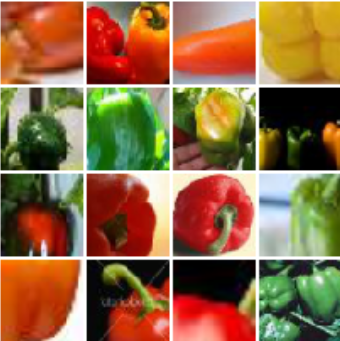}

\small \textbf{(b)} \dots but is often grouped with e.g., \texttt{bell pepper} when the image is a close-up view of its orange beak.
\caption{\label{fig:kingpenguin} Per-sample routing uncovers meaningful intra-class modes.
}
\end{wrapfigure}
Finally, we qualitatively analyze the experts' behavior:  
For each sample, we record which expert reaches minimal cross-entropy loss on that given sample. We then display the resulting class distribution across experts of the whole training set 
(see \hyperref[sec:specialization]{Appendix \ref{sec:specialization}}): 
The experts do end up specializing to specific subsets of the data, although, in contrast to hierarchical classification, many classes are still clearly split across several experts. 
Furthermore, most of these specialization patterns are consistent across the number of experts. 
Finally, while some experts clearly account for more classes than others, no expert is ever fully inactive. 
Looking more closely at the routed samples, we also see that the router uncovers natural intra-class variations, as illustrated for the class \texttt{king penguin} in \hyperref[fig:kingpenguin]{Figure \ref{fig:kingpenguin}}: The class \texttt{king penguin} is split across (i) close-up images where the orange beak is very visible, and end up being mapped to the same expert as oranges, bellpepper etc,  which it is often confused with, and (ii) far away images which are instead grouped with other animal classes.

\vspace{-0.15cm}
\section{Conclusions}\label{sec:conclusion}
\vspace{-0.2cm}
In this work, we revisit the single-gate Mixture of Experts (MoE) for convolutional architectures.
Specifically, we augment MoE with a novel ensembling scheme and a simple asynchronous and stable training pipeline leveraging a clustering-based initialization. 
Our model consistently reaches higher accuracy than hierarchical classifiers and a 1-expert ensembling baseline, revealing the benefits of training  specialized experts with per-sample routing.
Moreover, maintaining the base model as an independent branch allows us to further save computations at inference time using a simple threshold-based conditional rule.

Finally, our model is competitive with recent multi-layer MoE dynamic routing works, despite a smaller number of routers and experts, and  provides a more lightweight and stable training pipeline.
In the future, we plan to further improve our model's training efficiency by investigating different sampling strategies based on the gate outputs.

\bibliography{egbib}

\newpage

\begin{center}
\LARGE Supplemental Material to BMVC2022

\centering  ``Revisiting Single-gated Mixtures of Experts''
\end{center}
\vspace{0.2cm}

 \setcounter{section}{0}  
\section{Datasets Detailed Description}
\label{app:datasets}
We perform experiments on three datasets which we describe below:
\begin{itemize}
\item CIFAR-100~\cite{Krizhevsky2009LearningML} (32x32 images, 50k training samples, 10k test samples, 100 classes) is a classic benchmark for small-scale image classification.
\item tiny-ImageNet~\cite{tinyimagenet} (64x64 images, 100k training samples, 10k test samples, 200 classes) is a downscaled subset of ImageNet. It is significantly more challenging than CIFAR-100, which, combined with its reasonable scale, allows us to perform comprehensive ablation experiments in this study. 
\item ILSVRC2012~\cite{Russakovsky2015ImageNetLS} (224x224 images, 1.28M training samples, 50k test samples, 1000 classes) is the de-facto benchmark for large-scale classification.
\end{itemize}

\section{Training hyperparameters}
\label{app:training-details}
In this section, we report the hyperparameters we used to train our baselines and run our experiments.

\paragraph{CIFAR-100.} We follow the training pipeline of \cite{Devries2017ImprovedRO} to train our CIFAR-100 baselines.
Each model is trained for \textbf{200} epochs with an initial learning of \textbf{0.1}; The learning rate is decayed by a factor of \textbf{5} at epoch \textbf{60}, \textbf{120} and \textbf{160}. The model is trained with SGD with a momentum of \textbf{0.9}. Finally, we use a batch size of\textbf{ 128}. 
For training the experts, we use the same hyperparameters, but train them with batch size \textbf{512} (4 times fewer training iterations), starting from pretrained weights from the base models.

\paragraph{tiny-ImageNet.} We follow the training pipeline of \cite{Li2021BoostingDV} to train our tiny-ImageNet baselines. Each model is trained for \textbf{400} epochs with an initial learning of \textbf{0.2}; The learning rate is decayed by a factor of \textbf{10} at epoch \textbf{200} and \textbf{300}. The model is trained with SGD with a momentum of \textbf{0.9}. Finally, we use a batch size of \textbf{256}. 
For training the experts, we use the same hyperparameters, but only train for \textbf{100} epochs and the same batch size of 256, starting from pretrained weights from the base models.

\paragraph{ImageNet.} We use the torchvision default pretrained models, whose training hyperparameters details can be found in \cite{torchparams}. 
We train all our experts with the same hyperparameters except \textbf{(i)} We use cosine learning rate decay and \textbf{(ii)} we use fewer training iterations (\textbf{45} epochs with batch size \textbf{2048} for our experts, versus \textbf{90} epochs with batch size \textbf{32} in the original ResNet18, and \textbf{600 }epochs with batch size \textbf{128} in the original MobileNetv3-small).
\clearpage

\section{Qualitative Analysis of Experts specialization}
\label{sec:specialization}
\begin{figure}[ht]
\centering
\includegraphics[trim={19.5cm 0 0 0}, clip, width=0.45\textwidth, ]{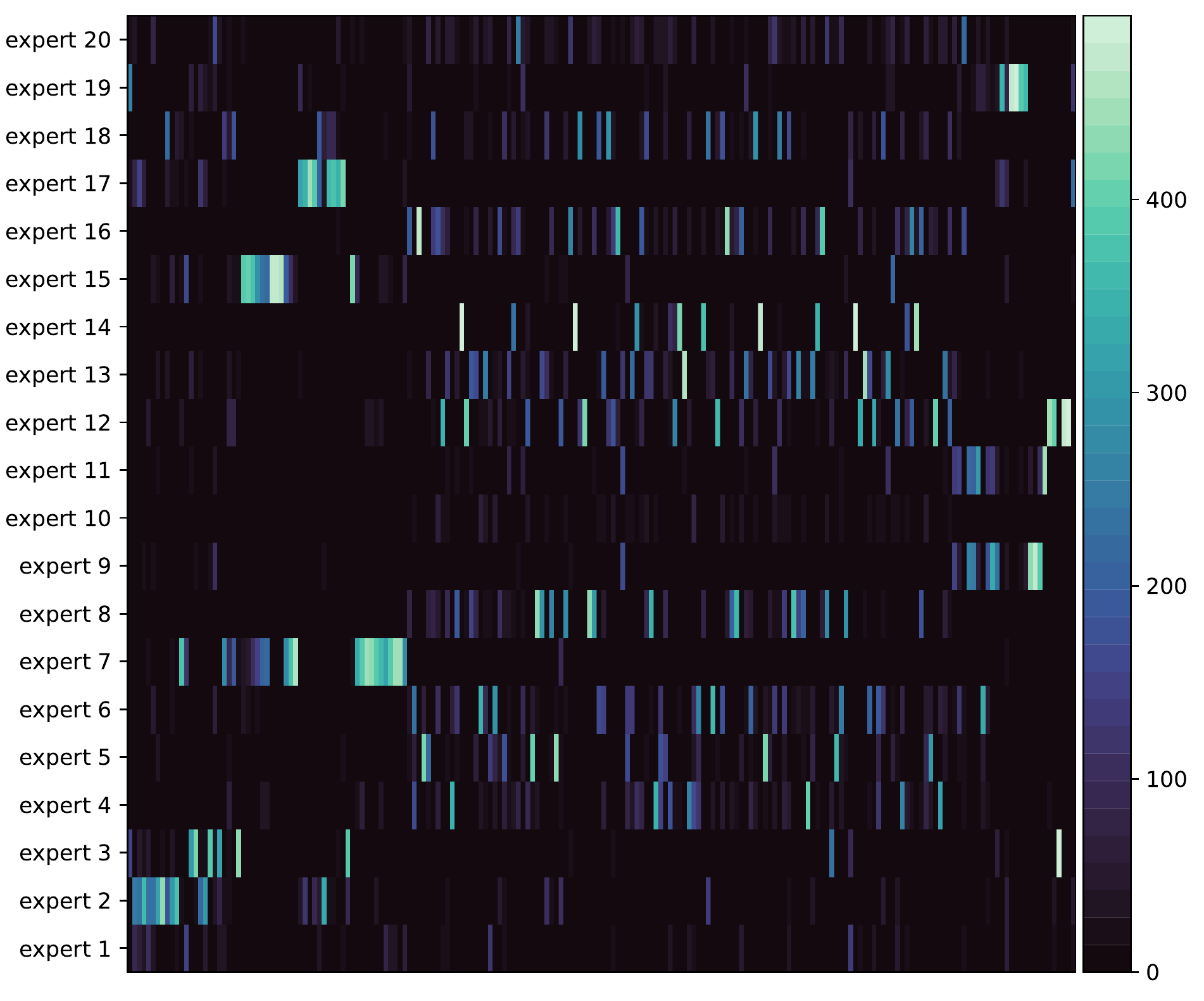}
\includegraphics[trim={19.5cm 0 0 0}, clip, width=0.45\textwidth, ]{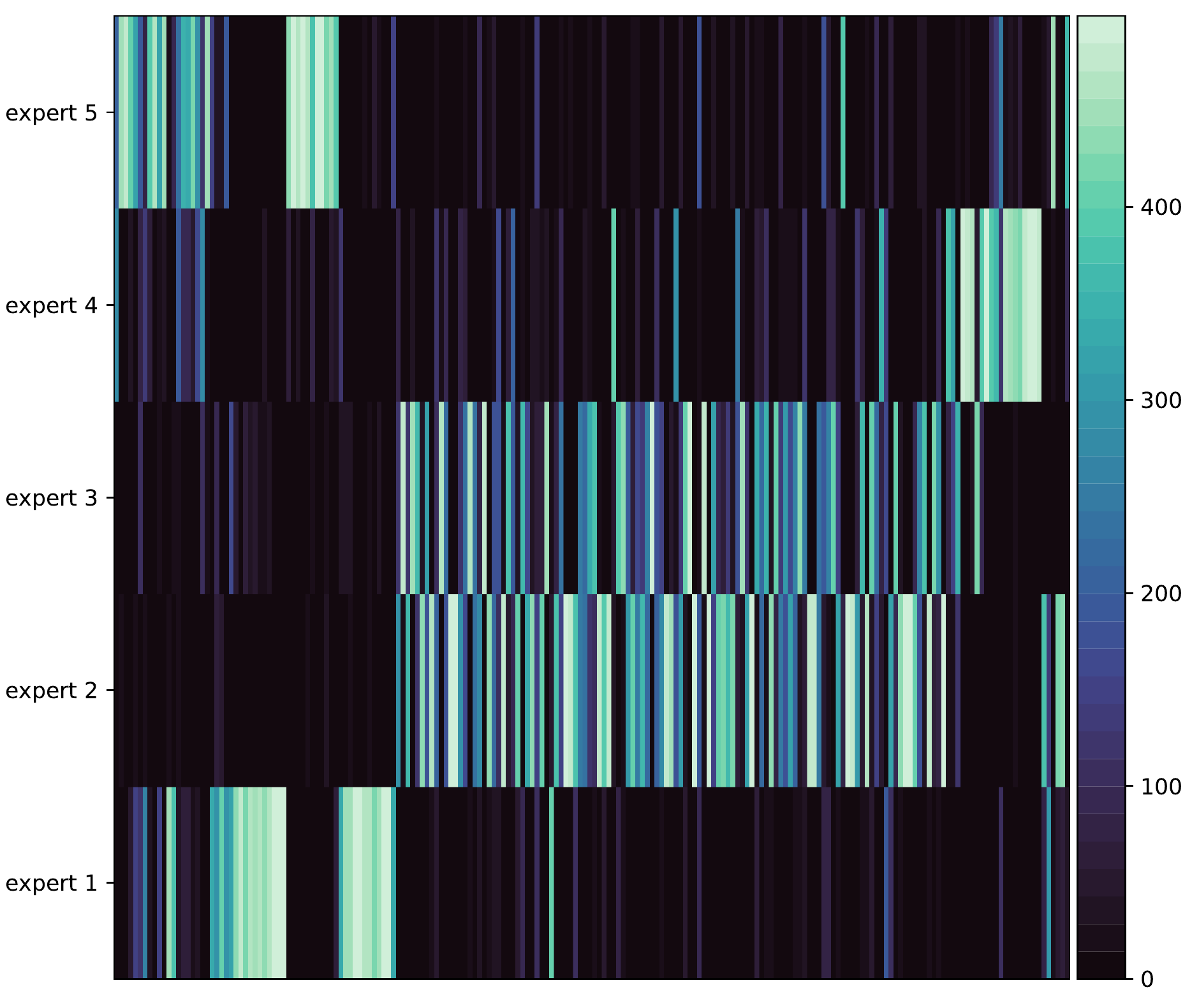}
\caption{\label{fig:specialization}Specialization pattern of the trained experts for 20 experts (\textit{left}) and 5 experts (\textit{right}) on the tiny-ImageNet dataset. The x-axis represents the 200 classes, and the y-axis represents the experts. The cells' colors represent the number of training samples of said class and for which said expert yields the lowest cross-entropy loss}
\end{figure}

In this section, we report on a few qualitative results to highlight how the experts specialize during training. 
\textbf{First}, we perform a simple analysis on trained experts: 
For each sample, we record which expert reaches minimal cross-entropy loss on that given sample. We then display the resulting class distribution across experts of the whole training set. 
We report the results in \hyperref[fig:specialization]{Figure \ref{fig:specialization}}: 
We observe that the experts do end up specializing to specific subsets of the data, although, in contrast to hierarchical classification, many classes are still clearly split across several experts. 
Furthermore, most of these specialization patterns are consistent across the number of experts. 
Finally, while some experts clearly account for more classes than others, no expert is ever fully inactive. 

\begin{figure}[!ht]
\centering
\includegraphics[width=0.48\textwidth]{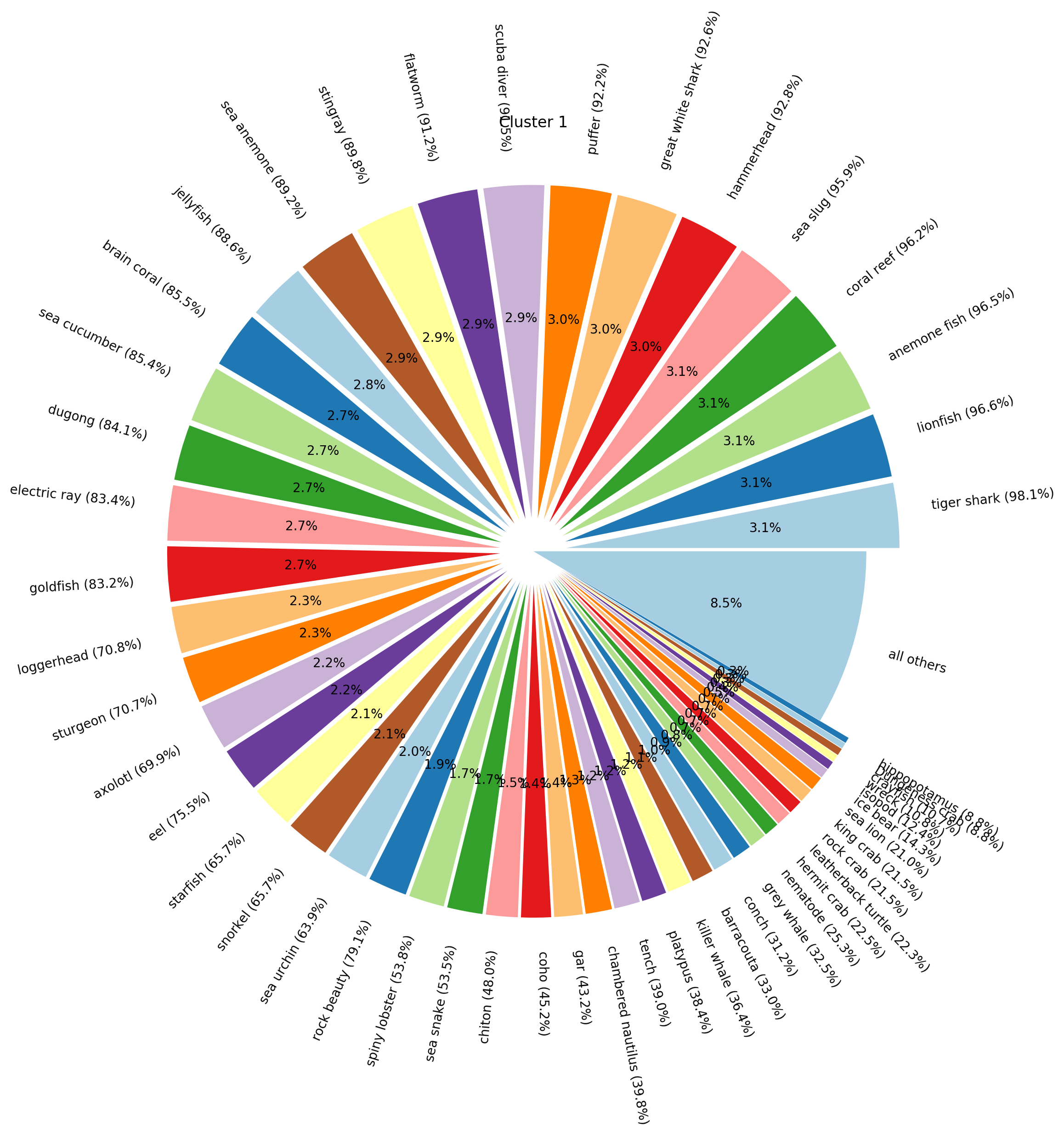}
\includegraphics[width=0.48\textwidth]{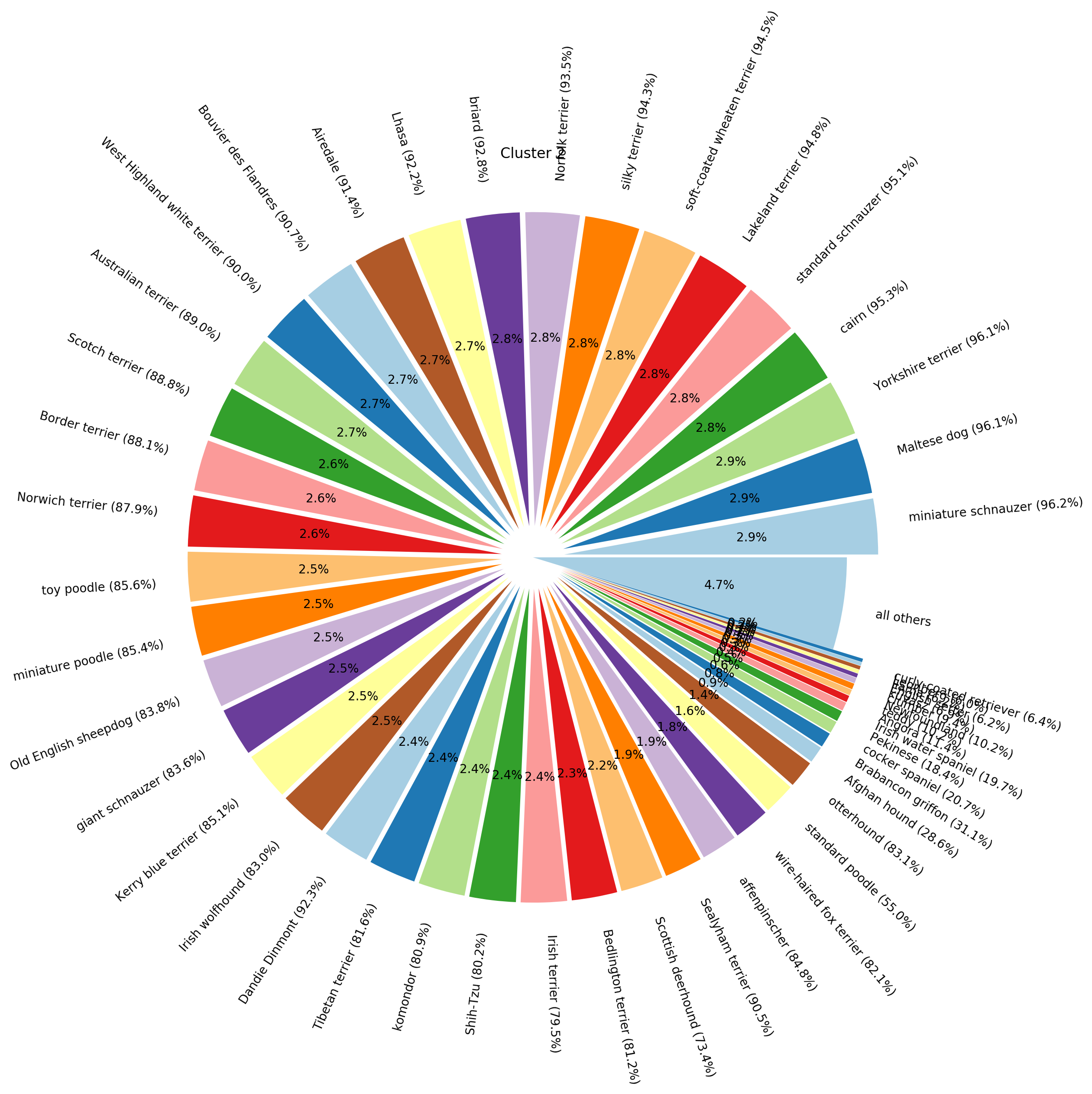}

\textbf{(a)} Example of two self-contained / highly specialized experts

\includegraphics[width=0.48\textwidth]{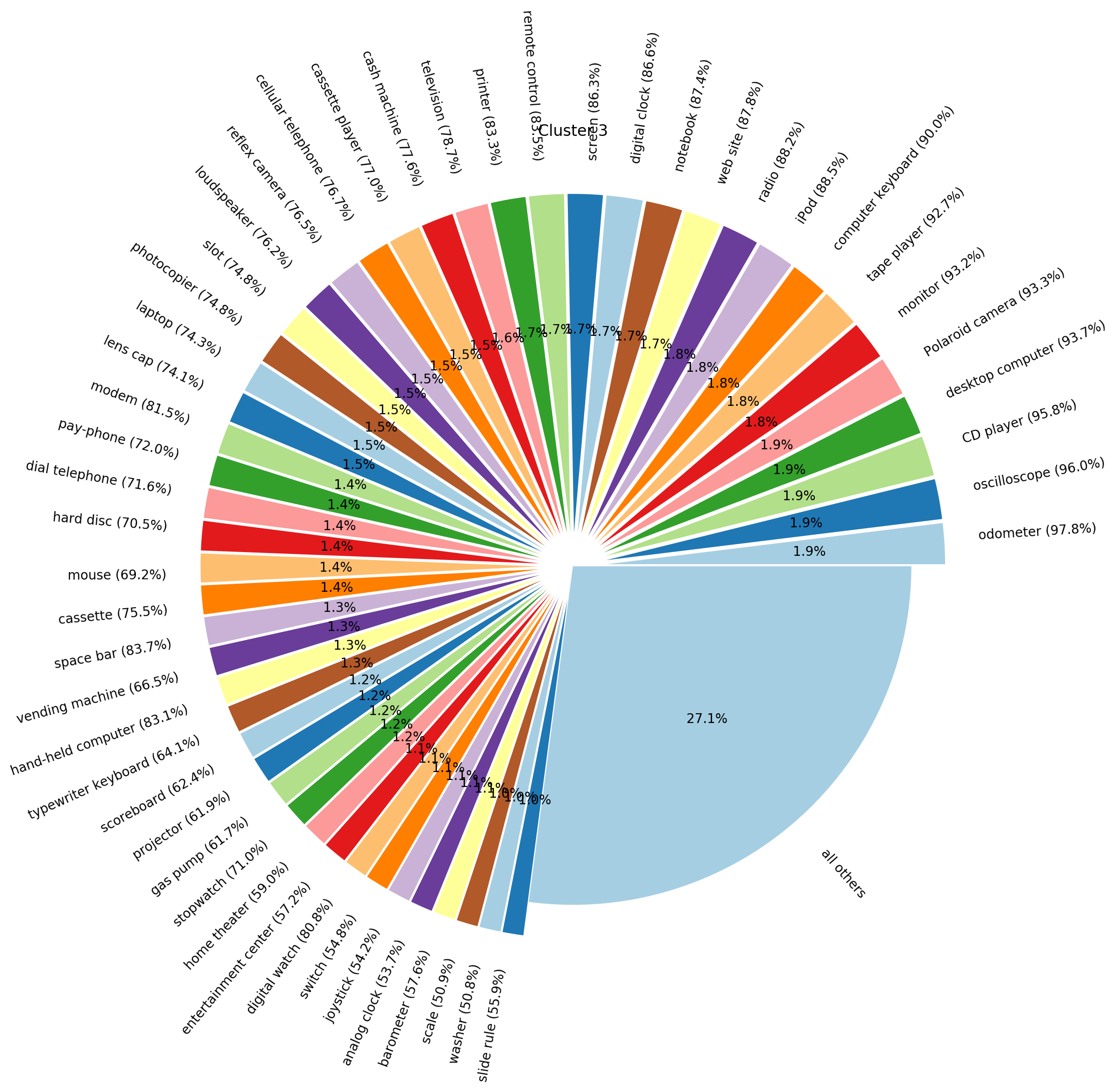}
\includegraphics[width=0.48\textwidth]{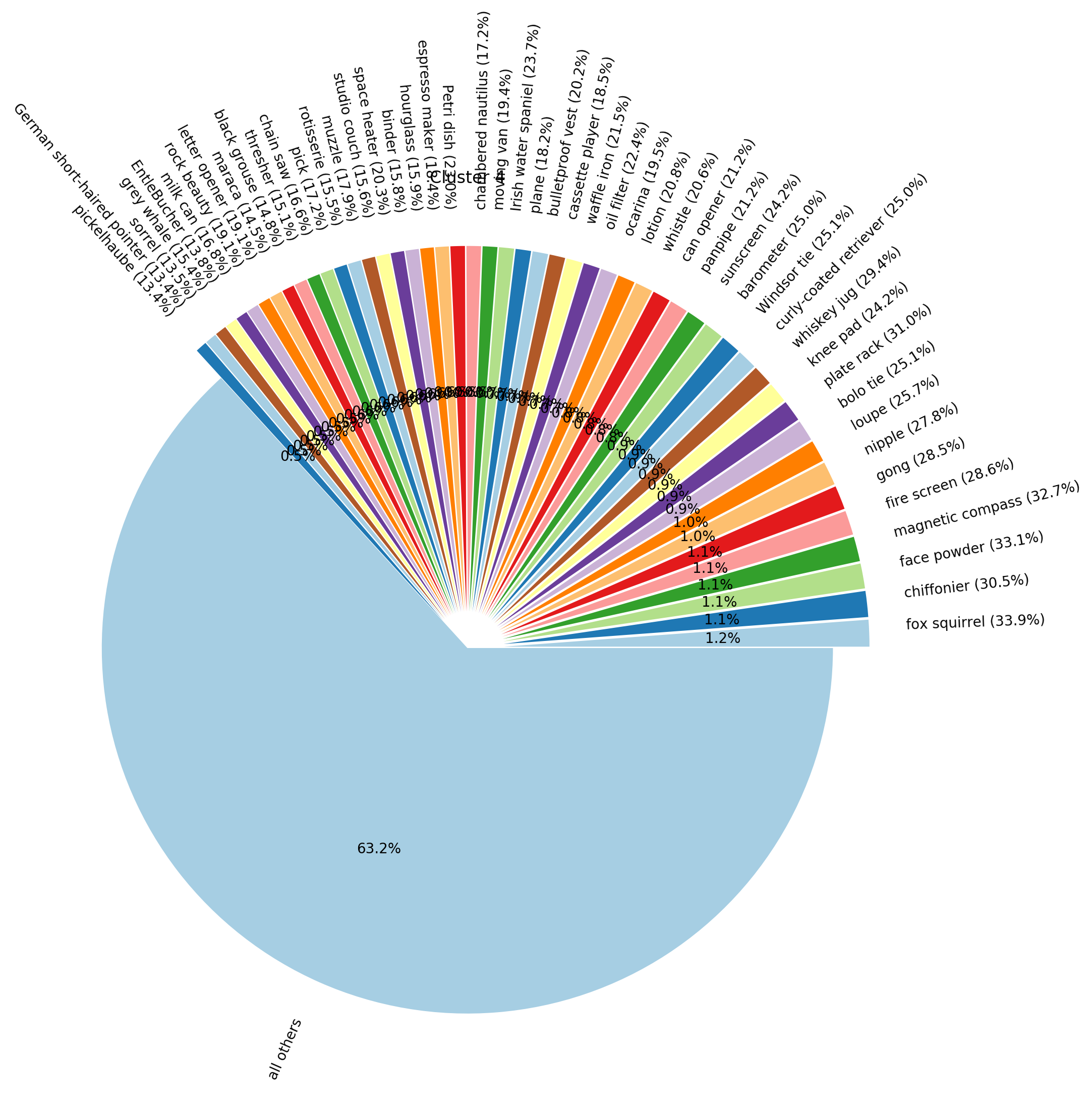}

\textbf{(b) }Example of complex experts with much higher class variation

\caption{\label{fig:kmeans}Visualization of some sample-to-expert mappings  from the precomputed initial gate $g_0$ on ImageNet for 20 experts and the ResNet18 base model. Each pie chart represents an expert; For readability, only the slices of the top-k classes are displayed individually. 
The percentage \textit{inside} a slice is the percentage of this class  among the samples mapped to this expert. 
Conversely, the percentage next to the class label is the percentage among all samples of the class (across all experts): A high percentage means that the class is almost entirely contained in one cluster. Best seen zoomed in.
}
\end{figure}

\textbf{Second}, we visualize the initial sample-to-expert assignment from the gate $g_0$,  following the $K$-means clustering step on the base model. 
We report some of these visualizations on ImageNet in \hyperref[fig:kmeans]{Figure \ref{fig:kmeans}}. 
The initial gate does find meaningful groupings in the dataset, following the base model's pretrained embeddings; Furthermore, the mapping does not exactly  respect the dataset's labels as most classes are distributed across more than one expert.
Finally, we observe that there are different levels of "density" across the clusters: E.g. the 9-th expert is very self-contained and contains almost entirely all dog breeds. 
In contrast the 6-th expert contains many more varied classes, and not always fully, mostly composed of common household objects.

\section{Analysis of the EM Training Scheme}
\label{sec:emappendix}
As described using the EM algorithm allows us to jointly train the gate alongside the experts, while preserving some training stability. 
In particular, a determining factor is $N_E$, the number of E steps update. The higher this number is, the more often the gate is updated based on experts feedback. Our asynchronous training pipeline (Algorithm `) can be seen as a special case $N_E = 0$. 

Given a fixed number of total epochs, $n_e$ we  vary the number of EM iterations, $N_E$. For our proposed training pipelines ($N_E = 0$) this means that the experts are trained for $n_e$ epochs with the fixed initial gate $g_0$. 
For $N_E > 0$, the experts are also trained for $n_e$ epochs, but every $\frac{n_e}{N_E + 1}$ epoch, the gate weights are updated based on the current experts' performance using the E step formula in \textbf{Equation 2}. 

\begin{figure}[h]
\begin{minipage}[c]{0.36\textwidth}
\centering
\resizebox{\textwidth}{!}{%
\begin{tabular}{|c|c|c|}
\hline
\multirow{2}{*}{$N_E$} & top-1-acc & top-1 acc \\
 & \scriptsize (w/o ensembling) & \scriptsize(w/ ensembling) \\
 \hline
0 \textbf{(ours)} & 63.11 & 65.72 \\
 \hline
 \hline
10 & 63.46 & 65.74 \\
 \hline
20 & 63.58 & 65.84 \\
 \hline
40 &  \textbf{64.48} & \textbf{66.15} \\
 \hline
\end{tabular}
}
\end{minipage}
\begin{minipage}[c]{0.61\textwidth}
\centering
\includegraphics[width=\textwidth]{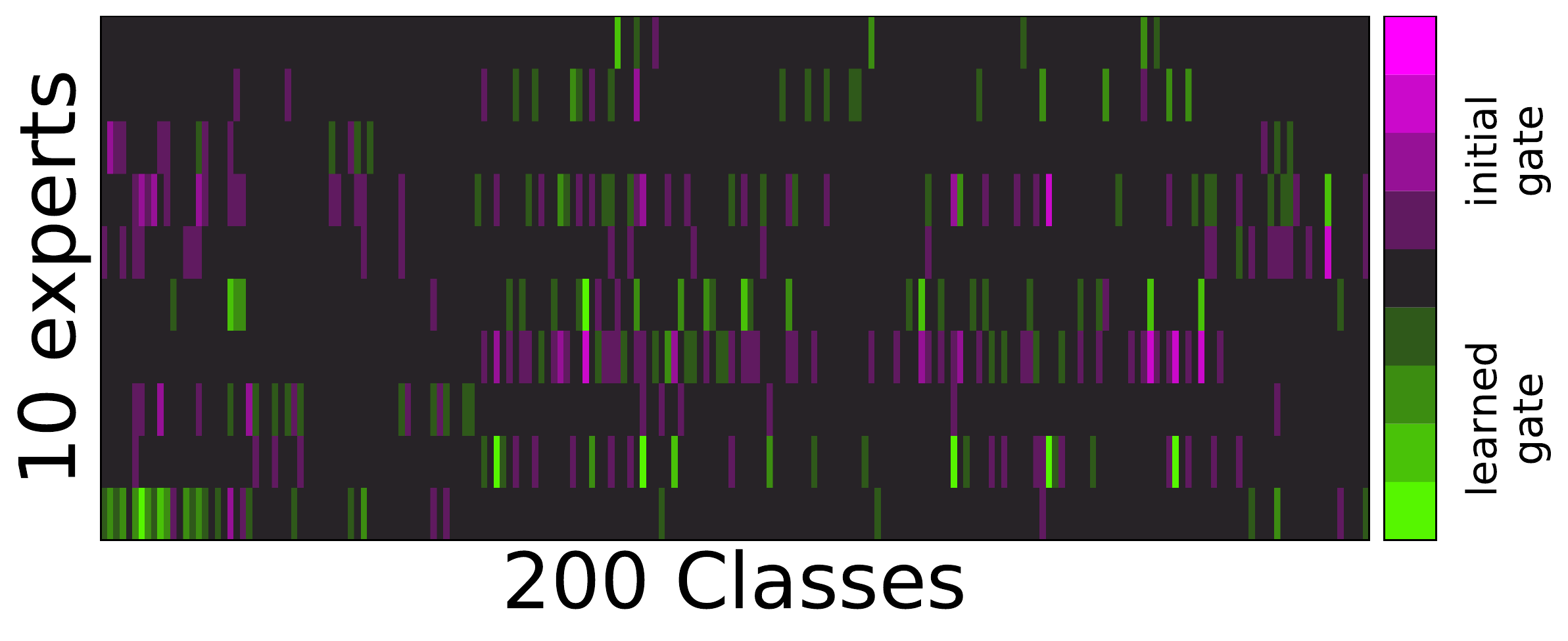}
\end{minipage}
\caption{\label{tab:em}(\emph{left}) Impact of increasing the number of E steps, $N_E$, when jointly training the gate and experts, using the  setup of  \hyperref[table:cifar]{Table \ref{table:cifar}} (tiny-ImageNet, 10 experts). The total training budget is the same for all runs, only the number of updates differ. (\emph{right}) We plot the heatmap of the difference between the class-to-expert assignments of the \textbf{\textcolor{RedViolet}{initial gate $g_0$}} and the \textbf{\textcolor{LimeGreen}{final trained gate}} for $N_E = 40$.}
\end{figure}

We report quantitative results on tiny-ImageNet in \hyperref[tab:em]{Figure \ref{tab:em} (left)}, and observe that joint training does result in more accurate models. However, this makes training cumbersome as it requires synchronization across experts every time the E step is computed, i.e., a total of $N_E$ times. 

For the case of $N_E = 40$, we further analyse the difference between the initial gate $g_0$ and the learned gate $g$ at the end of training, as shown in \hyperref[tab:em]{Figure \ref{tab:em}} : The two gates differ on roughly 14\% of the training dataset, and the assignment differences are not random but follow a specific pattern across classes.

We also detail the five most consistent differences in assignment between these two gates below: The samples assigned to a different expert by the learned gate $g$ are semantically meaningful: $g$ often re-assigns the samples to an expert that already contained a large proportion of its ground-truth class. 
 
\begin{itemize}
    \item[\checkmark] \textbf{40} samples of the class \texttt{steel arch bridge} and \textbf{105} samples of the class \texttt{triumphal arch} originally mapped to expert \textbf{4} by $g_0$ are mapped to expert \textbf{5} by the learned gate instead.
    \item[\checkmark] \textbf{96} samples of the class \texttt{espresso} originally mapped to expert \textbf{6} by $g_0$ are mapped to expert \textbf{7} by the learned gate instead.
    \item[\checkmark] \textbf{81} samples of the class \texttt{European Fire Salamander} originally mapped to expert \textbf{8} by $g_0$ are mapped to expert \textbf{1} by the learned gate instead.
    \item[x] \textbf{59} samples of the class \texttt{bikini} originally mapped to expert \textbf{2} by $g_0$ are mapped to expert \textbf{9} by the learned gate instead.
\end{itemize}

For reference, and to better understand each expert's specialty, the most often assigned expert to each class by the original $g_0$ gate in this experiment was as follows:
\begin{itemize}
    \item \textbf{Expert 1:} European fire salamander, bullfrog, tailed frog, American alligator, boa constrictor, trilobite, scorpion, tarantula, centipede, brain coral, slug, sea slug, spiny lobster, dugong, cockroach, sea cucumber, snorkel, coral reef

\item \textbf{Expert 2:} academic gown, apron, bikini, bow tie, cardigan, Christmas stocking, fur coat, kimono, miniskirt, neck brace, plunger, poncho, potter's wheel, punching bag, sock, sombrero, sunglasses, swimming trunks, teddy, vestment, ice lolly

\item \textbf{Expert 3:} goose, koala, Chihuahua, Yorkshire terrier, golden retriever, Labrador retriever, German shepherd, standard poodle, tabby, Persian cat, Egyptian cat, cougar, lion, brown bear, guinea pig, hog, ox, bison, bighorn, gazelle, Arabian camel, orangutan, chimpanzee, baboon, African elephant, lesser panda

\item \textbf{Expert 4:} abacus, altar, bannister, barbershop, brass, cash machine, chest, computer keyboard, confectionery, desk, dining table, freight car, organ, parking meter, pay-phone, refrigerator, rocking chair, scoreboard, sewing machine, space heater, turnstile, comic book

\item \textbf{Expert 5:} black stork, albatross, barn, beacon, birdhouse, cannon, cliff dwelling, crane, dam, flagpole, fountain, gondola, lifeboat, obelisk, picket fence, pole, projectile, steel arch bridge, suspension bridge, thatch, triumphal arch, viaduct, water tower, alp, cliff, lakeside, seashore

\item \textbf{Expert 6:} American lobster, frying pan, wok, plate, guacamole, ice cream, pretzel, mashed potato, cauliflower, bell pepper, orange, lemon, banana, pomegranate, meat loaf, pizza, potpie

\item \textbf{Expert 7:} goldfish, jellyfish, king penguin, backpack, barrel, bathtub, beaker, beer bottle, binoculars, broom, bucket, candle, CD player, chain, drumstick, dumbbell, gasmask, hourglass, iPod, lampshade, magnetic compass, nail, oboe, pill bottle, pop bottle, reel, remote control, sandal, stopwatch, syringe, teapot, water jug, wooden spoon, espresso

\item \textbf{Expert 8:} black widow, snail, ladybug, fly, bee, grasshopper, walking stick, mantis, dragonfly, monarch, sulphur butterfly, spider web, mushroom, acorn

\item \textbf{Expert 9:} basketball, butcher shop, jinrikisha, lawn mower, maypole, military uniform, rugby ball, torch, umbrella, volleyball

\item \textbf{Expert 10:} beach wagon, bullet train, convertible, go-kart, limousine, moving van, police van, school bus, sports car, tractor, trolleybus
\end{itemize}

\section{Derivation of the EM Algorithm}
\label{app:EM}
We denote by $X$ and $Y$ the random variables associated to the input data (images and associated class labels respectively).
We denote by $Z$ the hidden variable associated to the index of the expert that image $X$ is most likely associated to. 

We are interested in the two following probability densities:
\begin{itemize}
    \item $p(Z | X)$: The gate network routing a sample to an expert. This corresponds to the gate, $g(k | x)$, in  our model.
    \item $p(Y | X, Z = k)$: The probability distribution over class labels output by the $k$-th expert. This corresponds to the expert, $e_k(y | x)$, in our model.
\end{itemize}

We are interested in maximizing the total likelihood of the model, $p( y | x)$ using Expectation-Maximization (\textbf{EM}). 
We first derive the Evidence lower bound (ELBO) on $p(y | x)$ using the standard variational Bayesian framework; Introducing the variational distribution $q(Z | X, Y)$, we have:

\begin{align}
&KL(q(z| x, y) \| p(z | x, y)) = \mathbb E_q \log(q(z | x, y)) - \mathbb E_q \log p(z | x, y) \\
&= \mathbb E_q \log(q(z | x, y)) - \mathbb E_q \underbrace{\log p(z, y | x)}_{\log p(z | x) + \log p(y | z, x)} + \underbrace{\mathbb E_q \log p(y | x)}_{= \log p(y | x)} \\
&= KL(q(z|x, y) \| p(z | x)) - \mathbb E_q \log p(y | x, z) + \log p(y | x)
\end{align}

Reordering the terms around, we have:
\begin{align}
\log p(y | x) = \underbrace{\mathbb E_q \log p(y | x, z) - KL(q(z|x, y) \| p(z | x))}_{ELBO} + \underbrace{KL(q(z| x, y) \| p(z | x, y))}_{\geq 0}
\end{align}

The underlying idea of EM is a two step process in which we (i) minimize the difference between $\log p(y | x)$ and the ELBO (i.e., minimize the right-hand term in (11)) and (ii) maximize the ELBO (left-hand terms in (11)), usually  until convergence.
We then reiterate these two steps until satisfied. 

\vspace{0.25cm}

\textbf{E step: Minimize the difference between ELBO and the likelihood}

For this we only have to compute the posterior distribution, i.e., set $q$ to:

\begin{align}
q(z | x, y) &\leftarrow p(z | x, y) = \frac{p(z | x) p(y | z, x)}{p(y | x)} = \frac{p(z | x) p(y | z, x)}{\sum_z' p(z' | x) p(y | z', x)}\\
q(z | x, y)& \leftarrow \frac{g(z | x) e_z(y | x)}{\sum_z' g(z' | x) e_{z'}(y | x)}
\end{align}

\textbf{M - step: Maximize ELBO with the fixed q}
\begin{align}
ELBO &= \mathbb E_{z \sim q(z | x, y)} \log p(y | x, z) - KL(q(z|x, y) \| p(z | x))
\end{align}

Using our model's notations (and a sum over the experts  rather than an expectation), this means we have to maximize:
\begin{align}
 \sum_z \left[ q(z | x, y) \log e_z(y | x) - KL(q(z|x, y) \| g(z | x)) \right]
\end{align}

Thus we exactly recover \textbf{Equations 2} and \textbf{3} from the main text

\paragraph{Equivalence to backpropagation. }
The separation of the E and M step is crucial for training stability: In the M step, the updates for the parameters of the gating function and the experts become decoupled entirely, hence can be trained independently. 
In fact, if we perform the E and M steps on the same batch of data, then we have by definition that $q(z | x, y) = p(z | x, y)$ on the current batch. 
Therefore, the right-hand KL divergence term in \textbf{Equation 11} is equal to 0, and the ELBO is equal to the total log-likelihood.
Therefore, the EM algorithm simply becomes equivalent to directly minimizing the total log-likelihood $\log p(y | x)$ via backpropagation.

In other words, performing the E and M steps simultaneously is equivalent to directly training the gate and experts jointly with standard backpropagation. 
However, we observed many training instabilities using direct backpropagation (e.g. gate collapse), which also seems to be the case in the literature as most works introduce ad-hoc losses (e.g. balancing loss term) when implementing joint training via backpropagation.

Intuitively, this means that EM essentially implements a  delay in the parameter updates: We estimate the probability of the data-points belonging to each expert in the E step, and then we  update the gate and expert functions following that estimate for a few epochs.
The algorithm thus incurs one extra hyperparameter on top of standard backpropagation; That is, how many times we go through the E step to update the expert assignments estimates (we denote this hyperparamter as $N_E$ in the main text).

\section{Ablation: Expert Entry Layer}
\label{sec:expertentry}
In the main paper, we always set the number of early layers shared by the experts to be about half the architecture (i.e. 3 layers for ResNet-based models, and 8 for MobileNetv3-small). 

\begin{figure}[!ht]
\begin{minipage}[c]{0.49\textwidth}
\centering
\includegraphics[width=\textwidth]{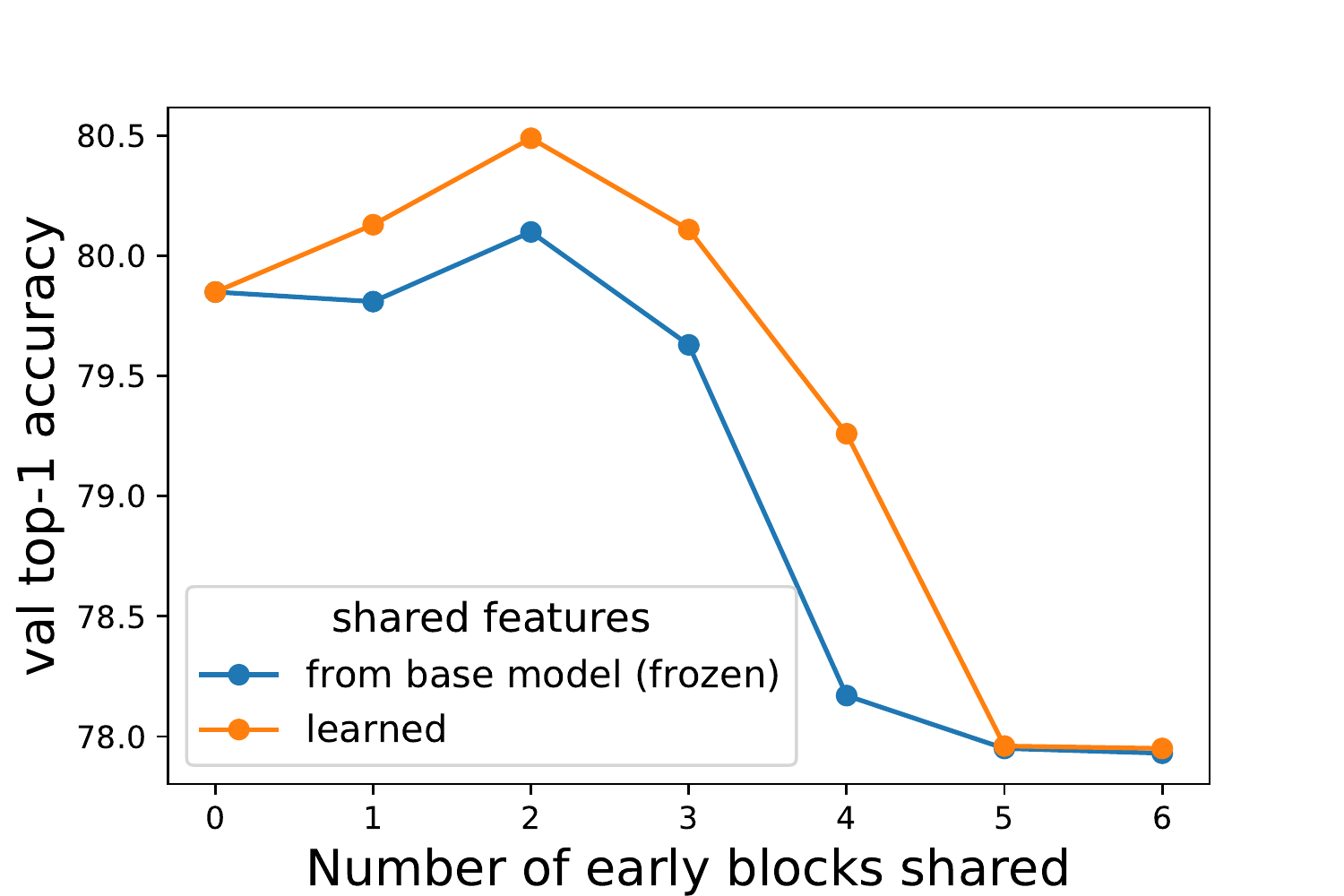}

(a) CIFAR-100
\end{minipage}
\begin{minipage}[c]{0.49\textwidth}
\centering
\includegraphics[width=\textwidth]{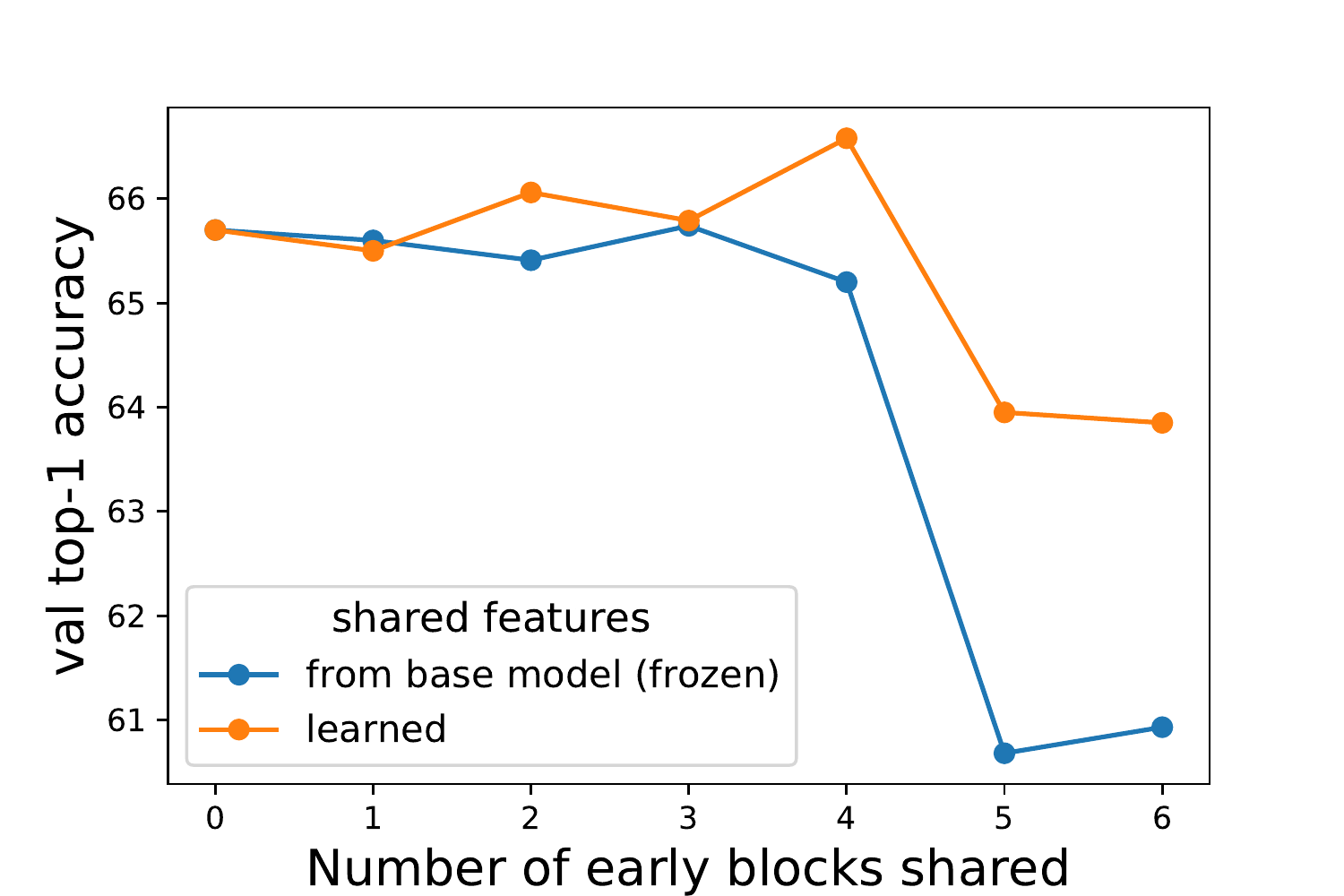}

(b) tiny-ImageNet
\end{minipage}
\caption{Validation accuracy when varying the number of early layers shared by the experts. All experiments were performed using 10 experts in the \texttt{tr18-tr18} configuration on CIFAR-100 (\textit{left}) and tiny-ImageNet (\textit{right}).\label{fig:layers}}
\end{figure}

Performing a full ablation study on this parameter shows that it could be better optimized for further accuracy gain: 
For instance in \hyperref[fig:layers]{Figure \ref{fig:layers}}, we observe that sharing 2 layers instead of 3 yields higher final accuracies on CIFAR-100. 
In both CIFAR-100 and tiny-ImageNet, we also see that there is a clear drop when the experts are reduced to only being linear layers. 
In general, we observe that the early features from the tiny-ImageNet base model are useful to the expert until a higher depth than the CIFAR ones. 
Automatically tuning this parameter for a given dataset and architecture is a  research direction we consider for future work. 

\section{Ablation: $\gamma$ Hyperparameter}
\label{app:gamma}
\begin{figure}[!ht]
    \centering
    \includegraphics[width=0.68\textwidth]{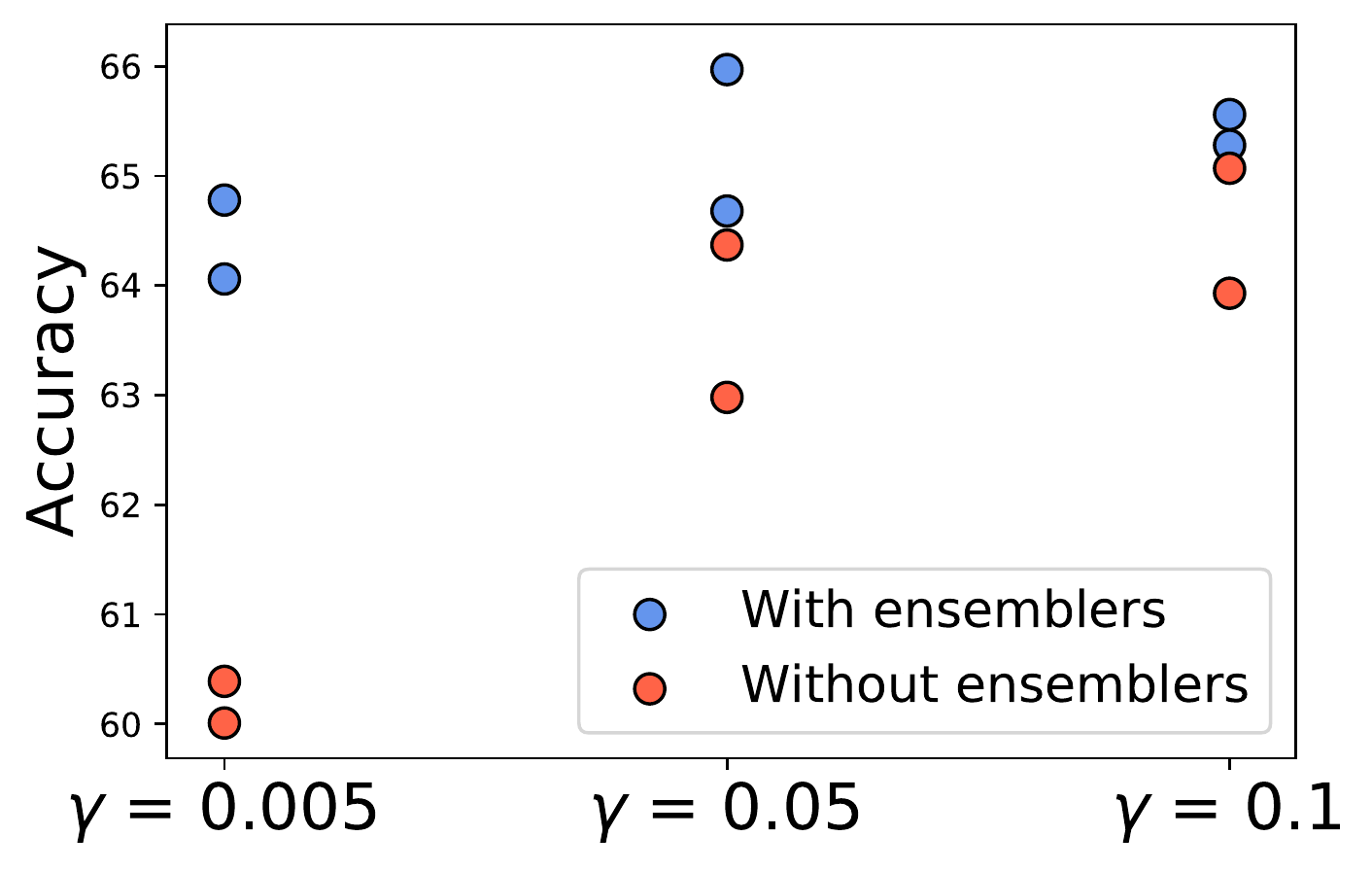}
    \caption{\label{fig:gamma} Effect of varying the $\gamma$ hyper-parameter (lower bound of the gate clipping operation $\Gamma$ when training the experts). Results are displayed for two random seeds, and obtained on tiny-ImageNet, for a \texttt{tr18-tr18} model with 10 experts.}
\end{figure}

In this section, we analyze the effect of the $\gamma$ hyperparameter that we use to smooth the gate weights during training the experts, as described in \textbf{Section 2.2}: 
$\gamma$ can be interpreted as a non-zero weight given to all "negative samples" (those which the gate does not map to the current expert) while training  an expert.
In \hyperref[fig:gamma]{Figure \ref{fig:gamma}}, we report accuracy results when varying $\gamma$.
We observe that too low values of $\gamma$ are detrimental to accuracy, both with and without ensemblers. 
However the model seems to be more robust to higher values of $\gamma$. In practice, we use a value $\gamma = 0.05$ for all our main  experiments.  

\section{Ablation: Early-exiting based on the gate or base model}
\label{app:earlyexitablation}
\begin{figure}[!ht]
\centering
\includegraphics[width=0.9\textwidth]{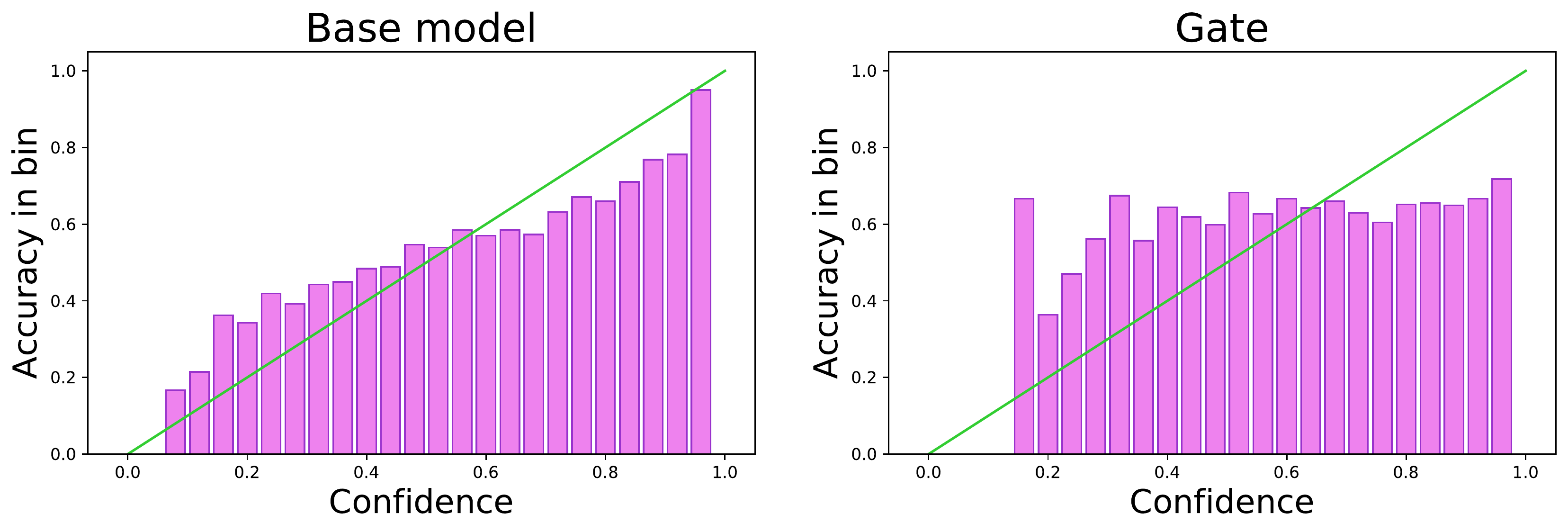}
\caption{\label{fig:reliability} Reliability pattern for \texttt{tr10} on tiny-ImageNet (20 \texttt{tr}-10 experts). The bins on the x-axis correspond to the confidence of the base model (\textit{left}) and the gate (\textit{right}).  The y-axis corresponds to the accuracy on the samples in each bin.}
\end{figure}

\paragraph{Thresholding based rule.} 
Our framework allows for two approaches to threshold-based early-exiting: (i) thresholding the gate's confidence; and (ii) thresholding the base model's confidence.
Thresholding the gate's confidence is based on the intuition that samples which are not routed confidently are also more likely to be classified wrong. 
On the other hand, base model thresholding directly builds on the intuition that the base model  predicts correct samples with high confidence while samples with lower confidence are more likely to be classified wrong. 
Based on the reliability diagrams in \hyperref[fig:reliability]{Figure \ref{fig:reliability}}, we can see that, while not perfectly calibrated, there is a strong correlation between confidence and accuracy for the base model, while the same is not the case for  the gate module. 
Therefore, we use the base model's confidence rather than the gate's as our guide for early-exiting in the main text and all experiments.

\begin{figure}[h!]
\centering
\includegraphics[width=0.95\textwidth]{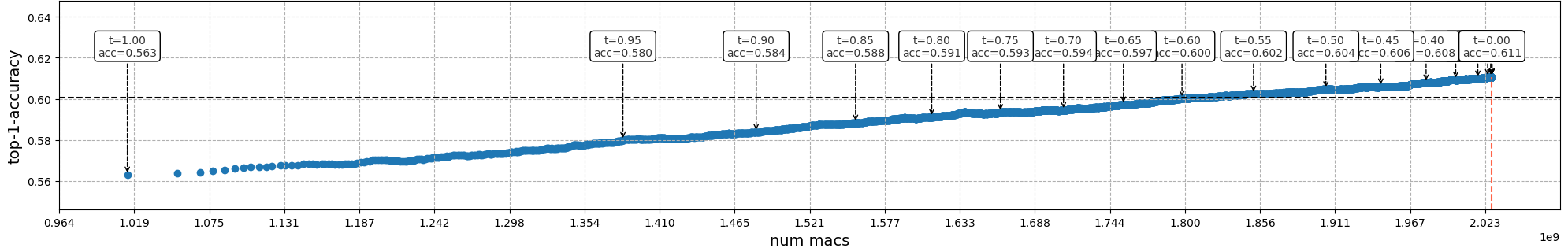}

\textbf{(a)} Gate's confidence thresholding

\includegraphics[width=0.95\textwidth]{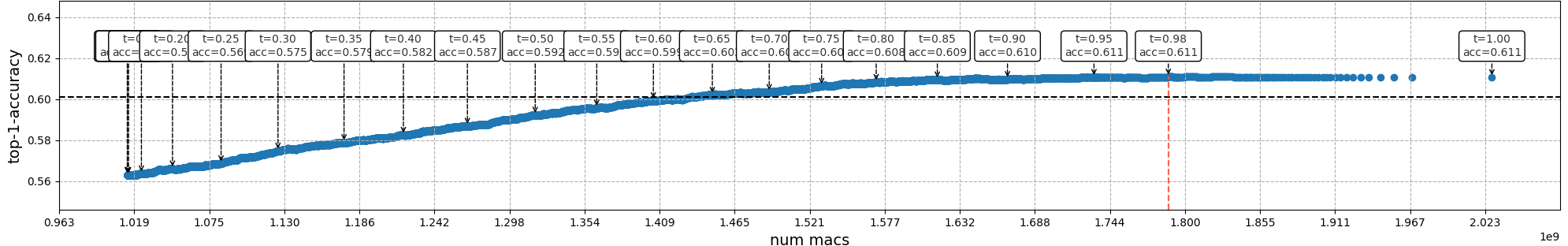}

\textbf{(b) } Base model's confidence thresholding

\caption{\label{fig:early-exiting-comparison} Comparison of early-exiting by thresholding either the  gate's confidence or the base model's confidence for \texttt{tr10}-base on tiny-ImageNet (20 \texttt{tr}-10 experts). 
The x-axis corresponds to number of MACs at different threshold values. The y-axis corresponds to accuracy of model with early-exiting. 
Dashed \textit{black} line corresponds to 1\% accuracy drop. Dashed \textit{red} line shows result for the best performing early-exiting  threshold. 
Boxes above the blue curve show the different threshold values along with accuracy for each  threshold.
}
\end{figure}

To confirm this observations further, we also plot early-exiting results for both gate's confidence thresholding and base model's confidence thresholding for one of our models in \hyperref[fig:early-exiting-comparison]{Figure \ref{fig:early-exiting-comparison}}. We observe that base model thresholding indeed yields higher accuracy vs early-exiting ratio trade-offs.

\paragraph{Selecting the best threshold.}
We further report results for the best performing threshold in each setting in \hyperref[tab:early-exiting-results]{Table \ref{tab:early-exiting-results}}. We present results for \texttt{tr10}-base on tiny-ImageNet (20 \texttt{tr}-34 experts) and \texttt{tr34}-base on tiny-ImageNet (20 \texttt{tr}-10 experts).
Looking at the whole range of thresholds allows us to find a model that fits the required computational budget. And selecting the best threshold by looking at the validation set can be seen as the upper bound. 
However, ideally, setting the optimal threshold properly and fairly would require an external validation holdout set. 
As an alternative, we use a subset of the training data to investigate whether we can set a threshold that generalizes well to inference time using training data only.
We report the corresponding results in \hyperref[tab:early-exiting-results]{Table \ref{tab:early-exiting-results}}.

\begin{table}[h!]
\centering
\resizebox{\textwidth}{!}{%
\begin{tabular}{|l|l|c|c|c|c|c|}
\hline
\multicolumn{1}{|c|}{method}                                & \multicolumn{1}{c|}{\begin{tabular}[c]{@{}c@{}}base\\ -\\ expert\\ -\\ num experts\end{tabular}} & \begin{tabular}[c]{@{}c@{}}baseline acc, \%\\ (no early-exiting)\end{tabular} & \begin{tabular}[c]{@{}c@{}}early-exit\\ ratio, \%\end{tabular} & \begin{tabular}[c]{@{}c@{}}top-1\\ acc, \%\end{tabular} & \begin{tabular}[c]{@{}c@{}}MACs \\ (x 1e9)\end{tabular} & \begin{tabular}[c]{@{}c@{}}\#params \\ (x 1e7)\end{tabular} \\ \hline
\multicolumn{1}{|c|}{\multirow{2}{*}{validation-selection}} & tr10-tr34-20                                                                                   & 66.22                                                                         & 20.11                                                          & 66.27                                                   & 4.7                                                     & 2.2                                                         \\ \cline{2-7} 
\multicolumn{1}{|c|}{}                                      & tr34-tr10-20                                                                                   & 66.36                                                                         & 29.07                                                          & 66.36                                                   & 5.3                                                     & 2.5                                                         \\ \hline\hline
\multirow{2}{*}{subset-selection}                           & tr10-tr34-20                                                                                   & 66.22                                                                         & 37.40                                                          & 65.87                                                   & 3.9                                                     & 1.8                                                         \\ \cline{2-7} 
                                                            & tr34-tr10-20                                                                                   & 66.36                                                                         & 75.07                                                          & 65.18                                                   & 4.9                                                     & 2.3                                                         \\ \hline \hline
\multirow{2}{*}{gate-learning}                              & tr10-tr34-20                                                                                   & 66.22                                                                         & 10.64                                                          & 65.54                                                   & 5.1                                                     & 2.4                                                         \\ \cline{2-7} 
                                                            & tr34-tr10-20                                                                                   & 66.36                                                                         & 14.63                                                          & 65.61                                                   & 5.5                                                     & 2.5                                                         \\ \hline
\end{tabular}%
}
\caption{Best performing thresholds for different approaches to early exiting as discussed in \hyperref[app:earlyexitablation]{Appendix  \ref{app:earlyexitablation}}. \texttt{Validation-selection} stands for selecting optimal threshold on a validation set. \texttt{Subset-selection} stands for selection threshold on a subset of training data and using that for validation set. \texttt{Gate learning} stands for learning a separate gate for early exiting.}
\label{tab:early-exiting-results}
\end{table}

\paragraph{Learning the early-exiting gate.} Alternatively, we can try to \emph{train} the early-exiting behavior for a given early-exiting budget. 
This approach requires slightly changing model the architecture by adding another output to the gate $g$. 
We rewrite the early-exiting criterion to include this new output: 
\begin{align}
ee(x) &= \mathbbm{1}(arg\max_{k} g(k | x) = K + 1) \label{eq:ee-new}
\end{align}
where $\mathbbm{1}$ is the indicator function, $K$ is the number of experts, and the $K+1$ index corresponds to the newly added early-exiting output.
The drawback of such approach is that it requires to train  additional  parameters. However, since the rest of the network is kept fixed, training only requires a very small number of epochs and does not increase the  complexity of the entire approach much. 

Similar to previous works~\cite{Chen2020LearningTS,Scardapane2020DifferentiableBI}, we use a supervised approach to train the new gate.
However, unlike previous papers which use auxiliary classifiers to label each sample as being fit for early-exiting or not, we use a simple approach to label the data from the solution of an integer linear programming (\textbf{ILP}) problem:  
Specifically, we first fix an efficiency budget $\tau \in [0, 1]$, which is our target early exiting ratio on the training set. We then solve the optimal early exit assignment for budget $\tau$ on the training set by solving the following discrete optimization problem:

\begin{align*}
\text{maximize} \sum_{(x, y)} \left[ ee(x)\ \phi(y|x) + \left(1 - ee(x)\right) \ \sum_{k=1}^K g(k|x)\ e'_k(y|x) \right] \\
\end{align*}
subject to:
\begin{align*}
\frac{1}{N} \sum_x ee(x)  &= \tau \\
\forall x,\ ee(x) & \in \{0, 1\}
\end{align*}
where $N$ is the total number of samples, the assignment $ee(x) \in \{0, 1\}$ is 1 when sample $x$ should early exit, and 0 otherwise.
This is a simple ILP problem with binary variables, that directly solves the optimal assignment on the training set. Solving this problem yields binary labels $ee(x)$ which we then use to learn the early-exiting gate. 
We report accuracy results of the learned gate in  \hyperref[tab:early-exiting-results]{Table \ref{tab:early-exiting-results}}.
Overall, this approach seems to perform worse than selecting a simple threshold based on the training set, as the learned early exiting behavior does not generalize well at inference time`. 

\end{document}